\documentclass[letterpaper]{article} 
\usepackage{aaai24}  
\usepackage{times}  
\usepackage{helvet}  
\usepackage{courier}  
\usepackage[hyphens]{url}  
\usepackage{graphicx} 
\usepackage{natbib}  
\usepackage{caption} 
\usepackage{algorithm}
\usepackage{amsmath,amssymb,amsthm,bm}
\usepackage{algpseudocode}
\usepackage{multirow}
\usepackage{booktabs}
\usepackage[switch]{lineno} %
\usepackage{enumerate}
\usepackage{color}

\newtheorem{theorem}{Theorem}
\newtheorem{lemma}{Lemma}
\newtheorem{corollary}{Corollary}
\newtheorem{remark}{Remark}

\usepackage{newfloat}
\usepackage{listings}
\DeclareCaptionStyle{ruled}{labelfont=normalfont,labelsep=colon,strut=off} 
\floatstyle{ruled}
\newfloat{listing}{tb}{lst}{}
\floatname{listing}{Listing}
\title{Ahpatron: A New Budgeted Online Kernel Learning Machine\\
with Tighter Mistake Bound}
\author{
    Yun Liao,
    Junfan Li,
    Shizhong Liao\thanks{Corresponding author.},
    Qinghua Hu,
    Jianwu Dang
    }
\affiliations{

    College of Intelligence and Computing, Tianjin University, Tianjin 300350, China\\
    \{yliao,junfli,szliao,huqinghua\}@tju.edu.cn,
    jdang@jaist.ac.jp
}

\usepackage{bibentry}

\begin{document}

\maketitle

\begin{abstract}
  In this paper, we study the mistake bound of online kernel learning on a budget.
  We  propose a new  budgeted online kernel learning model,
  called Ahpatron,
  which significantly improves the mistake bound of previous work
  and resolves the open problem posed by Dekel, Shalev-Shwartz, and Singer (2005).
  We first present an aggressive variant of Perceptron,
  named AVP,
  a model without budget,
  which uses an active updating rule.
  Then we design a new budget maintenance mechanism,
  which removes  a half of examples,
  and projects the removed examples onto a hypothesis space spanned by the remaining examples.
  Ahpatron adopts the above mechanism to approximate AVP.
  Theoretical analyses prove that Ahpatron has tighter mistake bounds,
  and experimental results show that Ahpatron outperforms the state-of-the-art algorithms on the same
  or a smaller budget.
\end{abstract}

\section{Introduction}

    Online kernel methods are popular approaches to
    solving both online machine learning and offline machine learning problems
    \cite{Kivinen2001Online,Crammer2003Online,Lu2016Large,Koppel2019Parsimonious}.
    Let $(\mathbf{x}_t,y_t)$ be a sequence of examples,
    where $\mathbf{x}_t\in\mathbb{R}^d$ is an instance,
    and $y_t\in \{-1,1\}$ is its label,
    $t=1,2,\ldots,T$.
    Online kernel learning algorithms process the examples on the fly,
    and produce a sequence of hypotheses $\left\{f_t\right\}^{T+1}_{t=1}$
    from a reproducing kernel Hilbert space (RKHS) $\mathcal{H}$.
    At any round $t$,
    the algorithms maintain $f_t=\sum^{t-1}_{\tau=1}a_{\tau}\kappa\left(\mathbf{x}_{\tau},\cdot\right)$
    and thus must store $S_t=\left\{\left(\mathbf{x}_{\tau},y_{\tau}\right),a_{\tau}\neq 0,\tau\leq t-1\right\}$
    in memory.
    $S_t$ is called the \textit{active set} \cite{Dekel2005The}.
    The examples in the active set play a similar role to
    the support vectors in Support Vector Machine (SVM) \cite{vapnik1998statistical}.
    The size of $f_t$ is linear with respect to $t$,
    and the computational complexity
    (space complexity and per-round time complexity) is in $O(dt)$,
    which hinders the deployment of online kernel learning algorithms
    on computing devices with bounded memory resources.

    For bounded memories,
    many algorithms that only store $B\geq 1$ examples have been proposed
    \cite{Crammer2003Online,Weston2005Online,Dekel2005The,Cheng2006implicit,Cesa-Bianchi2006Tracking,Wang2012Breakingecond,Li2023Improved},
    where $B$ is the budget.
    Although the algorithms guarantee constant memory and per-round running time,
    it also poses a critical question:
    \textit{How does the prediction performance of the algorithms vary with the budget?}
    From the experimental perspective,
    many results have empirically showed that
    the larger the budget is, the better the algorithms will perform
    \cite{Dekel2005The,Zhao2012Fast,Wang2012Breakingecond,He2014Simple,Zhang2019Incremental}.
    From the theoretical perspective,
    the algorithms that only store $B$ examples
    have worse regret bounds or mistake bounds
    than those that store all of the examples
    \cite{Dekel2005The,Cesa-Bianchi2006Tracking,Orabona2008The,Zhao2012Fast,Wang2012Breakingecond,He2014Simple}.
    Let
    $M_T=\left\{t\in[T]:\mathbf{x}_t~\mathrm{is}~\mathrm{misclassfied}\right\}$,
    $\mathbb{H}=\left\{f\in\mathcal{H}:\Vert f\Vert_{\mathcal{H}}\leq U\right\}$,
    $U=\frac{\sqrt{B+1}}{4\sqrt{\ln{(B+1)}}}$
    \footnote{There is an open problem in \cite{Dekel2005The}:
    whether it is possible to remove the $\ln^{-\frac{1}{2}}{(B+1)}$ factor in $U$.
    We have solved the open problem in this paper.},
        and $\ell_{\mathrm{Hinge}}(\cdot,\cdot)$ be the hinge loss function.
    For any  $f\in\mathbb{H}$,
    the Forgetron algorithm \cite{Dekel2005The} enjoys a mistake bound as follows.
    \begin{equation}
    \label{eq:NeurIPS2022:mistake_bound:Forgetron}
    \begin{aligned}
        \vert M_T\vert &\leq 4L_T(f)+2\Vert f\Vert^2_{\mathcal{H}},\\
        L_T(f)&=\sum^T_{t=1}\ell_{\mathrm{Hinge}}(f(\mathbf{x}_t),y_t).\\
    \end{aligned}
    \end{equation}
    And given $f\in\mathbb{H}$,
    the RBP algorithm \cite{Cesa-Bianchi2006Tracking}
    enjoys an expected mistake bound as follows.
    \begin{equation}
    \label{eq:NeurIPS2022:mistake_bound:RBP}
    \mathbb{E}\left[\vert M_T\vert\right] \leq \gamma L_T(f)+\gamma UB+0.5\gamma U\sqrt{B}\ln{\left(0.5\gamma B^2\right)},
    \end{equation}
    where $U \leq\frac{\gamma-1}{\gamma+1}\sqrt{B}$
    \footnote{Although RBP removes the $\ln^{-\frac{1}{2}}{(B+1)}$ factor in $U$,
    the mistake bound is an expected bound that is weaker than the deterministic bound.
    Thus the problem is still open \cite{Dekel2008The}.}.
    The mistake bound is far from optimal since it becomes worse while $B$ increases.
    For instance,
    if $B=O\left(T\right)$, then $\mathbb{E}\left[\vert M_T\vert\right]=O(T)$.
    The POMDR algorithm \cite{Li2023Improved} enjoys a regret bound of
    $O\left(U\sqrt{\mathcal{A}_T T/B}+U\sqrt{\mathcal{A}_T}\right)$,
    where $\mathcal{A}_T$ is a complexity called kernel alignment.
    For any  $f\in\mathbb{H}$,
    the above regret bound naturally implies the following mistake bound.
    \begin{equation}
    \label{eq:AAAI2024:mistake_bound:POMDR}
        \vert M_T\vert =L_T(f) + O\left(U\sqrt{\mathcal{A}_T}+
        \frac{U}{\sqrt{B}}\sqrt{\mathcal{A}_TT}\right).
    \end{equation}
    The convergence rate of the mistake bound is $O\left(\frac{1}{\sqrt{B}}\right)$.
    The larger $B$ is, the smaller the mistake bound will be.

    In summary,
    previous work does not give a satisfied answer to the fundamental question.
    There are two reasons.
    (i) The mistake bounds in \eqref{eq:NeurIPS2022:mistake_bound:Forgetron}
    and \eqref{eq:NeurIPS2022:mistake_bound:RBP}
    do not give the convergence rates with respect to $B$.
    (ii) Although the mistake bound in \eqref{eq:AAAI2024:mistake_bound:POMDR}
    gives a convergence rate of $O\left(\frac{1}{\sqrt{B}}\right)$,
    the coefficient depends on $\sqrt{\mathcal{A}_TT}$ and is unsatisfied.
    It can be proved that $\mathcal{A}_T\geq \inf_{f\in \mathbb{H}}L_T(f)$
    in certain conditions.

    In this paper,
    we will propose a new algorithm, Ahpatron,
    and answer the question better.
    We first consider the case of no budget,
    and propose a variant of Perceptron \cite{Frank1958The}, called AVP,
    using a more active updating strategy.
    We prove that AVP improves the mistake bound of Perceptron.
    Then we propose Ahpatron that approximates AVP
    and enjoys a $L_T(f)+O\left(\frac{L_T(f)}{\sqrt{B}}\right)$ mistake bound that is better than all of previous results.
    Ahpatron uses a very simple but effective budget maintaining approach
    proposed in \cite{Li2023Improved}, i.e., removing a half of examples.
    The novelties of Ahpatron include
    a strategy that selects the removed examples
    and a projection scheme that keeps the information of the removed examples.

\subsection{Main Results}

    Our main results are summarized as follows.
\subsubsection{Mistake Bound of AVP}

    For any $f\in\mathcal{H}$,
    the mistake bounds of AVP and Perceptron are
    \begin{align*}
        \vert M_T\vert &\leq 2L_T(f) + \Vert f\Vert^2_{\mathcal{H}}+\Delta_T\\
        \intertext{and}
        \vert M_T\vert &\leq 2L_T(f)+ \Vert f\Vert^2_{\mathcal{H}}
    \end{align*}
    respectively, where $\Delta_T\leq 0$.
    AVP improves the mistake bound of Perceptron,
    and the improvement is non-trivial.

\subsubsection{Mistake Bound of Ahpatron}

    Given $f\in\mathbb{H}$,
    the mistake bound of Ahpatron is
    \begin{equation}
    \label{eq:ECML2023:mistake_budget_trade_off:H-Forgetron}
    \begin{split}
     &\vert M_T\vert \leq L_T(f) + \Delta_T+\\
     &\max\left\{\frac{12U}{\sqrt{B}}L_T(f),\frac{0.9U}{\sqrt{B}}L_T(f)+
    \frac{\sqrt{B}}{2U}\Vert f\Vert^2_{\mathcal{H}}\right\},
    \end{split}
    \end{equation}
    where $\Delta_T\leq 0$.
    We improve the result in \eqref{eq:NeurIPS2022:mistake_bound:Forgetron},
    since the coefficient on $L_T(f)$ can be smaller.
    We improve the result in \eqref{eq:NeurIPS2022:mistake_bound:RBP},
    since the dependence on $B$ is better.
    We also improve the result in
    \eqref{eq:AAAI2024:mistake_bound:POMDR},
    since $\min_{f\in \mathbb{H}}L_T(f) =O(\sqrt{\mathcal{A}_TT})$ in certain conditions.

\subsubsection{Resolving Open Problem}

    Let $U=\frac{\sqrt{B}}{4}$ in \eqref{eq:ECML2023:mistake_budget_trade_off:H-Forgetron}.
    Then the mistake bound of Ahpatron is
    $$
        \left\vert M_T \right\vert \leq
        \max\left\{4L_T(f),1.3L_T(f)+2\left\Vert f\right\Vert^2_{\mathcal{H}}\right\}+ \Delta_T.
    $$
    Here, we remove the $\ln^{-\frac{1}{2}}{(B+1)}$ factor in $U$,
    and improve the mistake bound in \eqref{eq:NeurIPS2022:mistake_bound:Forgetron}.
    Thus we resolve the open problem posed by \citet{Dekel2005The}.

\subsubsection{Refined Mistake Bound of Ahpatron}

    We further prove an algorithm-dependent mistake bound for Ahpatron,
    which can be much better than the mistake bound in
    \eqref{eq:ECML2023:mistake_budget_trade_off:H-Forgetron}.

    The mistake bound in \eqref{eq:ECML2023:mistake_budget_trade_off:H-Forgetron}
    depends on the selected kernel function
    and the structure of the examples via $L_T(f)$.
    In certain benign environments
    where we have $\min_{f\in\mathbb{H}}L_T(f) \ll T$,
    Ahpatron performs well using a small budget.
    For instance,
    the examples are linearly separable in the feature space induced by the kernel function.
    In the worst case,
    i.e, $\min_{f\in\mathbb{H}}L_T(f)\approx T$,
    our result still coincides with the result in \eqref{eq:AAAI2024:mistake_bound:POMDR}.

\subsection{Related Work}

    Instead of using a fixed budget,
    the Projectron and Projectron++ algorithm \cite{Orabona2008The}
    use the approximate linear dependence condition \cite{Engel2004The}
    to add the current example into the active set.
    However, the two algorithms
    can not precisely control the size of the active set,
    and may suffer a computational complexity in $O(dT^2)$.
    The mistake bounds of the two algorithms are also similar to the result in
    \eqref{eq:NeurIPS2022:mistake_bound:Forgetron}.

    Besides the budget maintaining technique,
    there are many other techniques
    that can keep constant memory for online kernel learning algorithms,
    such as random features \cite{Rahimi2007Random},
    Nystr\"{o}m approximation \cite{Williams2001}
    and matrix sketching \cite{Charikar2002Finding}.
    The FOGD algorithm \cite{Wang2013Large,Lu2016Large}
    uses random features to approximate kernel function
    and enjoys a $O\left(\sqrt{T}+\frac{T}{\sqrt{D}}\right)$ regret bound
    \footnote{The original regret bound is $O\left(\Vert f\Vert_1\sqrt{T}+\epsilon T\Vert f\Vert_1\right)$,
    where $\Vert f\Vert_1=\sum^T_{t=1}\vert a_t\vert$
    and $f=\sum^T_{t=1}a_t\kappa\left(\mathbf{x}_t,\cdot\right)$,
    and holds with probability $1-2^8\left(\sigma_p/\epsilon\right)^2\exp\left(-D\epsilon^2/(4d+8)\right)$.},
    where $D$ is the number of random features.
    The NOGD algorithm \cite{Wang2013Large,Lu2016Large} uses the Nystr\"{o}m technique,
    and enjoys a regret bound of $O\left(\sqrt{T}+\frac{T}{\sqrt{B}}\right)$.
    The two regret bounds imply a $L_T(f)+O\left(\sqrt{T}+\frac{T}{\sqrt{B}}\right)$ mistake bound,
    which is much worse than our mistake bound.
    The SkeGD algorithm \cite{Zhang2019Incremental} uses randomized sketching and
    enjoys a regret bound of $O\left(\sqrt{TB}\right)$
    under the assumption that the eigenvalues of the kernel matrix decay exponentially.
    Although the result is better by a constant $B$,
    it becomes worse in the case of $B=\Theta\left(T^\mu\right)$, $0<\mu < 1$.

\section{Problem Setting and Preliminaries}

\subsection{Notations}

    Let $\mathcal{I}_T:=\left\{\left(\mathbf{x}_t,y_t\right)\right\}_{t\in[T]}$ be a sequence of examples,
    where $\mathbf{x}_t\in\mathcal{X}\subseteq\mathbb{R}^d$ is an instance,
    $y_t\in \{-1,1\}$ is the label,
    and $[T] = \{1,\ldots,T\}$.
    Let $\kappa(\cdot,\cdot):\mathcal{X} \times \mathcal{X} \rightarrow \mathbb{R}$
    be a positive semidefinite kernel function
    and $\mathcal{H}$ be the associated RKHS,
    such that,
    (i) $\mathcal{H}=\overline{\mathrm{span}(\kappa(\mathbf{x}_t,\cdot)~\vert~t\in[T])}$,
    and (ii) for any $f\in\mathcal{H}, \mathbf{x}\in\mathcal{X}$,
    it must be $\langle f,\kappa(\mathbf{x},\cdot)\rangle_{\mathcal{H}}=f(\mathbf{x})$.
    Denote by $\langle\cdot,\cdot\rangle_{\mathcal{H}}$ the inner product in $\mathcal{H}$.
    For any $f\in\mathcal{H}$,
    there exists a weight vector ${\bf a}\in\mathbb{R}^T$ such that
    $f=\sum^T_{t=1}a_t\kappa(\mathbf{x}_t,\cdot)$.
    For another $g=\sum^T_{t=1}b_t\kappa(\mathbf{x}_t,\cdot)\in\mathcal{H}$,
    we define
    $$
        \langle f,g\rangle_{\mathcal{H}}=\sum^T_{t=1}\sum^T_{\tau=1}a_tb_\tau\kappa\left(\mathbf{x}_t,\mathbf{x}_\tau\right).
    $$
    The inner product induces the norm $\Vert f\Vert_{\mathcal{H}}=\sqrt{\langle f,f\rangle_{\mathcal{H}}}$.
    We further assume that
    $\max_{\mathbf{x}\in\mathcal{X}}\kappa(\mathbf{x},\mathbf{x}) \leq 1$.
    Let $\mathcal{X}$ be bounded.
    Usual kernel functions,
    such as polynomial kernels and radial basis kernels, satisfy the assumption.
    Our results are suitable for any $\kappa$ such that $\max_{\mathbf{x}\in\mathcal{X}}\kappa(\mathbf{x},\mathbf{x})$ is bounded.
    Denote by $\ell_{\mathrm{Hinge}}(u,y)=\max\{0,1-uy\}$ and
    $L_T(f)=\sum^T_{t=1}\ell_{\mathrm{Hinge}}\left(f\left({\bf x}_t\right),y_t\right)$, $f\in\mathcal{H}$.

\subsection{Budgeted Online Kernel Learning}

    The protocol of online kernel learning
    can be defined as a game between a learner and an adversary.
    At any round $t$,
    the adversary sends an instance $\mathbf{x}_t\in\mathcal{X}$.
    Then the learner selects a hypothesis $f_t\in\mathcal{H}$
    and makes a prediction $\hat{y}_t=\mathrm{sign}(f_t(\mathbf{x}_t))$.
    After that the adversary gives the label $y_t$.
    The game proceeds to the next round.
    We rewrite $M_T=\{t\in[T]: \hat{y}_t\neq y_t\}$.
    The learner aims to minimize $\vert M_T\vert$.
    For any $f\in\mathcal{H}$,
    we use an upper bound on the mistakes to measure the performance of the learner,
    $$
    \vert M_T\vert \leq h\left(\mathcal{I}_T,f\right),
    $$
    where $h(\mathcal{I}_T,f)$ depends on $\mathcal{I}_T$ and $f$.

    At any round $t$,
    $f_t=\sum^{t-1}_{\tau=1}a_{\tau}\kappa\left(\mathbf{x}_{\tau},\cdot\right)$.
    To store $f_t$ in memory,
    the learner must store
    $S_t=\left\{\left(\mathbf{x}_{\tau},y_{\tau}\right), a_{\tau}\neq 0, \tau \leq t-1\right\}$.
    The memory cost may be unbounded.
    To keep the memory bounded,
    we would limit $\vert S_t\vert\leq B$.
    We call $B$ the budget.
    The budget also weakens the performance of the learner.
    It is impossible to provide a non-trivial mistake bound
    for any $f\in\mathcal{H}$ without additional assumptions on the problem,
    so we maintain a competitive hypothesis space \cite{Dekel2005The},
    $$
        \mathbb{H}=\left\{f\in\mathcal{H}: \Vert f\Vert_{\mathcal{H}}\leq U\right\},
    $$
    where $U$ is a constant.

    The state-of-the-art mistake bound is \cite{Li2023Improved}
    $$
    h\left(\mathcal{I}_T,f\right)=L_T(f)+ O\left(U \sqrt{\frac{\mathcal{A}_T T} {B}}\right).
    $$
    Our goal is to design an algorithm that achieves
    $$
    h\left(\mathcal{I}_T,f\right) = L_T(f) + O\left(\frac{U}{\sqrt{B}}L_T(f)\right).
    $$
    Such an upper bound is tighter than previous results.

\subsection{Perceptron}
\label{sec:ECML2022:analysis_of_OKS}

    We first consider online kernel learning without budget.
    A classical algorithm with non-trivial mistake bounds is Perceptron \cite{Frank1958The}.
    At any round $t$,
    Perceptron predicts $\hat{y}_t=\mathrm{sign}(f_t({\bf x}_t))$.
    The key is the update rule,
    $$
        f_{t+1}=f_t+y_t\kappa\left({\bf x}_t,\cdot\right)\cdot\mathbb{I}_{\hat{y}_t\neq y_t}.
    $$
    \begin{theorem}[\citet{Dekel2005The}]
    \label{thm:ECML2023:mistake_bound_Perceptron}
        For any sequence of examples $\mathcal{I}_T$ and for any $f\in\mathcal{H}$,
        the mistake bound of Perceptron satisfies
        $$
        \vert M_T\vert \leq 2L_T(f) +\Vert f\Vert^2_{\mathcal{H}}.
        $$
    \end{theorem}
    \begin{theorem}[\citet{Shai2007Online}]
    \label{thm:ECML2023:mistake_bound_Perceptron_ada_learning_rate}
        For any $\mathcal{I}_T$ and for any $f\in\mathcal{H}$,
        the mistake bound of Perceptron satisfies
        $$
        \vert M_T\vert \leq L_T(f) +\Vert f\Vert_{\mathcal{H}}\sqrt{L_T(f)}+\Vert f\Vert^2_{\mathcal{H}}.
        $$
    \end{theorem}

    If $L_T(f)\leq \Vert f\Vert^2_{\mathcal{H}}$,
    then the mistake bound in Theorem \ref{thm:ECML2023:mistake_bound_Perceptron}
    is better than that in Theorem \ref{thm:ECML2023:mistake_bound_Perceptron_ada_learning_rate}.
    Perceptron stores all of the misclassified examples.
    Its space complexity is in $O\left(d\vert M_T\vert\right)$,
    which would be $O(dT)$ in the worst case.
    Many algorithms approximate Perceptron
    via maintaining a fixed budget,
    such as Forgetron \cite{Dekel2005The} and RBP \cite{Cesa-Bianchi2006Tracking}.

\section{AVP}

    Perceptron uses a passive updating rule.
    A more active updating rule is that
    $f_t$ makes a prediction at a low confidence,
    i.e., $y_tf_t({\bf x}_t)< \beta_t$, $\beta_t\in[0,1]$.
    The updating rule is adopted by $\mathrm{ALMA}_p$ \cite{Gentile2001A}.
    If $\beta_t=0$, then it follows Perceptron.
    If $\beta_t=1$,
    it follows Projectron++ \cite{Orabona2008The} and
    the gradient descent family of algorithms \cite{Zhao2012Fast,Lu2016Large,Zhang2019Incremental}.
    However,
    no mistake bound shows that
    the active updating rule is better than that of Perceptron.
    There are two technical challenges,
    \begin{enumerate}[\textbf{C}~1]
      \item how to set the value of $\beta_t$;
      \item how to give a tighter analysis.
    \end{enumerate}
    We will carefully design $\beta_t$,
    and give a novel and solid theoretical analysis of mistake bounds.

    Let $\beta_t=1-\varepsilon_t$.
    We redefine the updating rule as follows.
    \begin{align*}
        f_{t+1}
        &=\mathrm{Proj}_{\mathbb{H}}\left(f_t+\lambda_t\cdot y_t\kappa\left({\bf x}_t,\cdot\right)\cdot
        \mathbb{I}_{y_t\cdot f_t(\mathbf{x}_t)<1-\varepsilon_t}\right),\\
        \mathrm{Proj}_{\mathbb{H}}(g)
        &=\min\left\{1,U\cdot \Vert g\Vert^{-1}_{\mathcal{H}}\right\}\cdot g,
    \end{align*}
    where $\varepsilon_t\in [0,1]$ and
    $\lambda_t$ is the learning rate.
    We will prove that setting $\frac{\lambda_t}{2} <\varepsilon_t<1$ will give smaller mistake bounds.
    We name this algorithm AVP (Aggressive Variant of Perceptron),
    and give the pseudo-code in Algorithm \ref{alg:ECML2023:Perceptron++}.

    \begin{algorithm}[!t]
        \caption{AVP}
        \footnotesize
        \label{alg:ECML2023:Perceptron++}
        \begin{algorithmic}[1]
        \Require {$\{\lambda_t,\varepsilon_t\}^T_{t=1},U$}
        \Ensure  $f_1=0$, $S_1=\emptyset$\;
        \For{$t=1$,\ldots,$T$}
            \State Receive ${\bf x}_t$;
            \State Compute $\hat{y}_t=\mathrm{sign}\left(f_t({\bf x}_t)\right)$;
            \If{$y_t\cdot f_t({\bf x}_t) < 1-\varepsilon_t$}
                \State Update $f_{t+1}=\mathrm{Proj}_{\mathbb{H}}\left(f_{t}+\lambda_t y_t\kappa\left({\bf x}_t,\cdot\right)\right)$
                \State $S_{t+1}=S_t\cup\left\{\left({\bf x}_t,y_t\right)\right\}$
            \EndIf
        \EndFor
        \end{algorithmic}
    \end{algorithm}

    \begin{theorem}
    \label{thm:ECML2023:Perceptron++}
        Let
        \begin{align*}
            M'_T &=\left\{t\in[T]: y_tf_t\left({\bf x}_t\right)\leq 0 \right\},\\
            N_T  &=\left\{t\in[T]: 0<y_t f_t\left({\bf x}_t\right) < 1-\varepsilon_t \right\}.\\
        \end{align*}
        Set
        $U >0$,
        $\lambda_t< 2$,
        $\delta_t=\Vert f_t-f\Vert^2_{\mathcal{H}} - \Vert f_{t+1}-f\Vert^2_{\mathcal{H}}$,
        and
        \begin{equation}
        \label{eq:AAAI24:value_of_varepsilon}
           \frac{\lambda_t}{2} < \varepsilon_t < 1.
        \end{equation}
        For any $f\in\mathbb{H}$,
        the mistake bound of AVP satisfies
        \begin{align*}
          &\left\vert M'_T\right\vert -L_T(f)\\
          \leq
          &\sum_{t\in M'_T\cup N_T}\frac{\delta_t}{2\lambda_t}
            +\sum_{t\in M'_T}\frac{\lambda_t}{2}
            +\sum_{t\in N_T}\left(\frac{\lambda_t}{2}-\varepsilon_t\right).
        \end{align*}
    \end{theorem}

    Note that $U=+\infty$ implies $\mathbb{H}=\mathcal{H}$.
    Recalling the definition of $M_T$,
    we have $M'_T=M_T\cup\{t\in[T]:\hat{y}_t=y_t,f_t({\bf x}_t)=0\}$.
    Thus $\vert M_T\vert \leq \vert M'_T\vert$.

    Eq. \eqref{eq:AAAI24:value_of_varepsilon} solves the first challenge \textbf{C}~1.
    To rise to the second challenge \textbf{C}~2,
    our theoretical analysis makes use of the gap between mistake bounds and cumulative hinge losses,
    which is given as follows.
    \begin{align*}
        &\sum_{t\in M'_T\cup N_T}\ell_{\mathrm{Hinge}}\left(f_t\left({\bf x}_t\right),y_t\right)-\vert M'_T\vert\\
        \geq& \sum_{t\in N_T}1-y_tf_t({\bf x}_t)\\
        \geq& \sum_{t\in N_T}\varepsilon_t.
    \end{align*}
    We will use the non-positive term $-\sum_{t\in N_T}\varepsilon_t$
    to counteract the increase of $\sum_{t\in N_T}\frac{\lambda_t}{2}$ on the upper bound.
    This non-positive term naturally improves the mistake bound of Perceptron
    and solves \textbf{C}~2.

    \begin{corollary}
    \label{coro:ECML2023:perceptron++:constant_learnining_rate}
        Let $U=+\infty$, $\lambda_t=1$, $\varepsilon_t=\varepsilon\in(\frac{1}{2},1)$.
        For any $ f\in\mathcal{H}$,
        the mistake bound of AVP satisfies
        $$
            \vert M_T\vert
            \leq
            2L_T(f) +
            \Vert f\Vert^2_{\mathcal{H}} +
            \left(1-2\varepsilon\right)\vert N_T\vert.
        $$
    \end{corollary}

    Note that $\left(1-2\varepsilon\right)\vert N_T\vert\leq 0$.
    Compared with Theorem \ref{thm:ECML2023:mistake_bound_Perceptron},
    AVP improves the mistake bound of Perceptron.
    It is worth mentioning that $\vert N_T\vert$ depends on $\varepsilon$.
    It seems that setting $\varepsilon=1$ minimizes $\left(1-2\varepsilon\right)\vert N_T\vert$.
    However,
    if $\varepsilon=1$, then $\vert N_T\vert=0$.
    It is hard to give the exact value of $\vert N_T\vert$.
    The only information is that $\vert N_T\vert \geq 0$
    for all $\varepsilon\in[0,1)$.
    According to the upper bound,
    it is better to set $\frac{1}{2}<\varepsilon<1$.
    In this case, $\vert N_T\vert> 0$
    unless all instances are classified with a high confidence.
    We will empirically verify that $\vert N_T\vert\gg 1$.

    \begin{corollary}
    \label{coro:ECML2023:ALMA++:time_learnining_rate}
        Let $U<+\infty$, $\lambda_t
        =\frac{U}{\sqrt{U^2+\sum^t_{\tau=1}\mathbb{I}_{y_{\tau}f_{\tau}({\bf x}_{\tau})\leq 0}}}$,
        $t\geq 1$ and
        $\frac{\lambda_t}{2} < \varepsilon_t < 1$.
        For any $ f\in\mathbb{H}$,
        the mistake bound of AVP satisfies
        \begin{align*}
            \vert M_T\vert
            \leq &\max\left\{L_T(f)+ 2 U^2 + \Delta_T, 0\right\} +9U^2+\\
                 &3U\sqrt{\max\left\{L_T(f)+2U^2+\Delta_T,0\right\}},
        \end{align*}
        where $\Delta_T=\sum_{t\in N_T}\left(\frac{\lambda_t}{2}-\varepsilon_t\right)$.
    \end{corollary}

    Compared with Theorem \ref{thm:ECML2023:mistake_bound_Perceptron_ada_learning_rate},
    the non-positive term $\Delta_T$ makes that AVP improves the mistake bound of Perceptron.
    AVP stores $\vert M'_T\vert+\vert N_T\vert$ instances
    in memory.
    In the worst case,
    the memory cost of AVP is also in $O(dT)$.
    In the next section,
    we will approximate AVP by limiting $\vert S_t\vert\leq B$.
    \begin{remark}
        The updating rule of AVP is similar to $\mathrm{ALMA}_2$ \cite{Gentile2001A}.
        The essential differences are the values of $\beta_t$ and $\lambda_t$.
        $\mathrm{ALMA}_2$ sets
        $$
            \beta_t=\frac{(1-\alpha)B}{\Vert{\bf x}_t\Vert^{-1}_2\sqrt{k}},
            \lambda_t=\frac{Ck^{-\frac{1}{2}}}{\Vert{\bf x}_t\Vert_2},
            k=1+\sum^{t-1}_{\tau=1}\mathbb{I}_{y_\tau f_\tau({\bf x}_\tau)\leq \beta_{\tau}},
        $$
        where $\alpha$, $B$ and $C$ are tunable parameters.
        In Corollary \ref{coro:ECML2023:ALMA++:time_learnining_rate},
        AVP sets $\beta_t=1-\varepsilon_t$ and
        $\lambda_t=\frac{U}{\sqrt{U^2+\sum^t_{\tau=1}\mathbb{I}_{y_{\tau}f_{\tau}({\bf x}_{\tau})\leq 0}}}$.

        The values of $\beta_t$ and $\lambda_t$ in $\mathrm{ALMA}_2$
        make its mistake bound be similar to
        Theorem \ref{thm:ECML2023:mistake_bound_Perceptron_ada_learning_rate}
        (see Theorem 3 in \citet{Gentile2001A}).
        Thus $\mathrm{ALMA}_2$ did not prove that more active updating rule is better than Perceptron.
        AVP also significantly improves the mistake bound of $\mathrm{ALMA}_2$.
    \end{remark}

\section{Ahpatron}

    Forgetron and RBP
    approximate Perceptron by limiting $\vert S_t\vert\leq B$.
    If $\vert S_t\vert=B$ and we will add $({\bf x}_t,y_t)$ into $S_t$,
    then Forgetron removes the oldest example, and
    RBP randomly removes an example.
    A recent algorithm POMDR \cite{Li2023Improved} adopts a very simple technique
    that removes a half of examples.
    We will approximate AVP by removing a half of examples.
    There are still two technical challenges.
    (1) How to select the examples that will be removed.
    (2) How to keep the information of the removed examples.
    We will propose a heuristic example selecting strategy
    and a novel projection scheme to address the two challenges.

    At any round $t$,
    if $\vert S_t\vert=B$ and $y_tf_t(\mathbf{x}_t) < 1- \varepsilon$,
    then we will remove $\frac{B}{2}$ examples from $S_t$.
    Let $f_t=\sum^B_{i=1}\alpha_{r_i}\kappa(\mathbf{x}_{r_i},\cdot)$,
    where $(\mathbf{x}_{r_i},y_{r_i})\in S_t$.
    Denote by $S_t=S_{t,1}\cup S_{t,2}$ satisfying
    (i) $S_{t,1}\cap S_{t,2}=\emptyset$;
    (ii) $\vert S_{t,1}\vert =\frac{B}{2}$;
    (iii) $\forall (\mathbf{x}_{r_i},y_{r_i})\in S_{t,2}$
    and $\forall (\mathbf{x}_{r_j},y_{r_j})\in S_{t,1}$,
    $\vert \alpha_{r_i}\vert\geq \vert \alpha_{r_j}\vert$.
    We rewrite $f_t = f_{t,1}+f_{t,2}$,
    where
    \begin{align*}
        f_{t,1}=&\sum_{\left(\mathbf{x}_{r_i},y_{r_i}\right)\in S_{t,1}}\alpha_{r_i}\kappa\left(\mathbf{x}_{r_i},\cdot\right),\\
        f_{t,2}=&\sum_{\left(\mathbf{x}_{r_i},y_{r_i}\right)\in S_{t,2}}\alpha_{r_i}\kappa\left(\mathbf{x}_{r_i},\cdot\right).
    \end{align*}
    We will remove the examples in $S_{t,1}$.
    If we directly reset $f_{t,1}=0$,
    then much of information may be lost.
    To keep as much information as possible,
    we propose a new projection scheme.
    The main idea is to project $f_{t,1}$ onto $\mathcal{H}_{t,2}=\overline{\mathrm{span}(\kappa(\mathbf{x},\cdot):(\mathbf{x},y)\in S_{t,2})}$.
    Let $\mathbb{H}_{t,2}=\{f\in\mathcal{H}_{t,2}:\Vert f\Vert_{\mathcal{H}}=c_tU\}$ where $c_t\in(0,1]$.
    $\forall f\in\mathcal{H}_{t,2}$,
    $f=\sum_{(\mathbf{x}_{r_i},y_{r_i})\in S_{t,2}}\theta_{r_i}\kappa(\mathbf{x}_{r_i},\cdot)$.
    The projection scheme is
    \begin{align}
        \hat{f}_{t,1} =& \mathop{\arg\min}_{f\in \mathcal{H}_{t,2}}\Vert f-f_{t,1}\Vert^2_{\mathcal{H}}
        +\eta\Vert {\bm \theta}\Vert^2_2,
        \label{alg:ICML2022:projection_step_1}\\
        \bar{f}_{t,2} =& \mathrm{Proj}_{\mathbb{H}_{t,2}}\left(f_{t,2}+\hat{f}_{t,1}\right),
        \label{alg:ICML2022:projection_step_2}
    \end{align}
    where $\eta\Vert {\bm \theta}\Vert^2_2$ is a regularization term and
    aims to make \eqref{alg:ICML2022:projection_step_1} solvable.
    We will explain more in \eqref{eq:ICML2022:theta:H-forgetron}.

    Let $f^{\top}_{t,1}=f_{t,1}-\hat{f}_{t,1}$
    and $f^\top_{t,2}=(f_{t,2}+\hat{f}_{t,1})-\bar{f}_{t,2}$.
    We further rewrite $f_{t}$ by
    \begin{align*}
        f_t = f_{t,2} + f_{t,1}
            =f_{t,2}+\hat{f}_{t,1}+f^{\top}_{t,1}
            =\bar{f}_{t,2}+f^{\top}_{t,2}+f^{\top}_{t,1}.
    \end{align*}
    For the sake of clarity, denote by $\bar{f}_t:=\bar{f}_{t,2}$.
    Then we execute
    \begin{equation}
    \label{eq:ICML2022:unified_half_perceptron:second_phase}
    \begin{split}
        \bar{f}_t=&f_t-(f^{\top}_{t,1}+f^{\top}_{t,2}),\\
        f_{t+1}=&\mathrm{Proj}_{\mathbb{H}}\left(\bar{f}_t+\lambda y_t\kappa(\mathbf{x}_t,\cdot)\right).
    \end{split}
    \end{equation}
    It is obvious that $(f^{\top}_{t,1}+f^\top_{t,2})$ is removed from $f_t$.
    Recalling that $\bar{f}_t$ is a combination of $\frac{B}{2}$ examples.
    Thus the first step in \eqref{eq:ICML2022:unified_half_perceptron:second_phase}
    will remove a half of examples from $S_t$.
    We name this algorithm Ahpatron (approximating aggressive Perceptron via halving and projecting)
    and give the pseudo-code in Algorithm \ref{alg:ICML2022:Half-Forgetron}.

    \begin{algorithm}[!t]
        \caption{\small{Ahpatron}}
        \footnotesize
        \label{alg:ICML2022:Half-Forgetron}
        \begin{algorithmic}[1]
        \Require{$B$, $U$, $\lambda$, $\varepsilon$, $\eta$, $\{c_t\}^T_{t=1}$}
        \State Initialize $f_0=0$, $S_1=\emptyset$
        \For{$t=1,\ldots,T$}
            \State Receive ${\bf x}_t$
            \State Compute $\hat{y}_t=\mathrm{sign}(f_t(\mathbf{x}_t))$
            \If {$y_tf_t(\mathbf{x}_t) < 1- \varepsilon$}
                \If{$\vert S_t\vert<B$}
                    \State Update $f_{t+1}=\mathrm{Proj}_{\mathbb{H}}\left(f_{t}+\lambda_t y_t\kappa\left({\bf x}_t,\cdot\right)\right)$
                    \State Update $S_{t+1}=S_t\cup\{(\mathbf{x}_t,y_t)\}$
                \Else
                    \State Divide $S_t=S_{t,1}\cup S_{t,2}$\;
                    \State Compute $\hat{f}_{t,1}, \bar{f}_{t,2}$
                    following \eqref{alg:ICML2022:projection_step_1} and \eqref{alg:ICML2022:projection_step_2}
                    \State Remove $S_{t,1}$ and update $f_{t+1}$
                    following \eqref{eq:ICML2022:unified_half_perceptron:second_phase}
                \EndIf
            \EndIf
        \EndFor
    \end{algorithmic}
    \end{algorithm}


    The projection \eqref{alg:ICML2022:projection_step_1} can be rewritten as follows,
    \begin{align*}
        \min_{{\bm \theta}\in \mathbb{R}^{\frac{B}{2}}}
        \left\{{\bm \theta}^\top(\mathbf{K}_2+\eta\mathbf{I}){\bm \theta}
        -2{\bm \theta}^\top\mathbf{K}_{21}{\bm \alpha}\right\}
        +{\bm \alpha}^\top\mathbf{K}_1{\bm \alpha},
    \end{align*}
    where ${\bm \alpha}=(\alpha_{r_i})_{(\mathbf{x}_{r_i},y_{r_i})\in S_{t,1}}$,
    $\mathbf{K}_2$ is the kernel matrix defined on $S_{t,2}$,
    $\mathbf{K}_1$ is the kernel matrix defined on $S_{t,1}$,
    $\mathbf{K}_{21}$ is the kernel matrix define on $S_2$ and $S_1$
    satisfying $\mathbf{K}_{21}[i,k]=\kappa(\mathbf{x}_{r_i},\mathbf{x}_{r_k})$,
    $(\mathbf{x}_{r_i},y_{r_i})\in S_{t,2}$ and $(\mathbf{x}_{r_k},y_{r_k})\in S_{t,1}$.
    The optimization problem has a closed-form solution,
    \begin{equation}
    \label{eq:ICML2022:theta:H-forgetron}
        {\bm \theta}^\ast= \left(\mathbf{K}_2+\eta\mathbf{I}\right)^{-1}\mathbf{K}_{21}{\bm \alpha},
    \end{equation}
    where $\eta>0$ aims to keep
    $\left(\mathbf{K}_2+\eta\mathbf{I}\right)^{-1}$ invertible.
    We further have
    \begin{align*}
    \hat{f}_{t,1}&=
    \sum_{(\mathbf{x}_{r_i},y_{r_i})\in S_{t,2}}\theta^\ast_{r_i}\kappa\left(\mathbf{x}_{r_i},\cdot\right),\\
    \bar{f}_t
    &=
    \frac{c_tU}{\Vert f_{t,2}+\hat{f}_{t,1}\Vert_{\mathcal{H}}}
        \sum_{\left(\mathbf{x}_{r_i},y_{r_i}\right)\in S_{t,2}}
        \left(\alpha_{r_i}+\theta^\ast_{r_i}\right)\kappa\left(\mathbf{x}_{r_i},\cdot\right).
    \end{align*}
    The time complexity of computing ${\bm \theta}^\ast$ is $O\left(B^3\right)$
    since there is a matrix inversion operation.
    Note that we solve ${\bm \theta}^\ast$ only if $\vert S_t\vert=B$,
    that is,
    the time interval between two continuous matrix inversion operations is at least $\frac{B}{2}$.
    Thus the average per-round time complexity is only $O\left(\max\{dB,B^2\}\right)$.
    The space complexity of Ahpatron is also $O\left(\max\{dB,B^2\}\right)$.

\section{Theoretical Analysis}

    Lemma \ref{lemma:ECML2023:number_of_removing} upper bounds
    the times of removing operation,
    which is critical to the mistake bound analysis.

    \begin{lemma}
    \label{lemma:ECML2023:number_of_removing}
        For any sequence $\mathcal{I}_T$,
        the times of removing operation executed by Ahpatron
        is at most
        $$
        \max\left\{\frac{2\left(\left\vert M'_T\right\vert+\left\vert N_T\right\vert\right)}{B}-1,0\right\},
        $$
        where $M'_T$ and $N_T$ are defined in Theorem \ref{thm:ECML2023:Perceptron++}.
    \end{lemma}

    \begin{theorem}[Mistake Bound]
    \label{thm:AAAI24:mistake_bound_of_H_Forgetron}
        Let $c_t=0.6$ for all $t\in[T]$,
        $\lambda=\frac{\sqrt{2}U}{\sqrt{B}}$, $B\geq 50$, $U\leq \frac{\sqrt{B}}{4}$
        and $\frac{3U}{\sqrt{B}}< \varepsilon <1$.
        Given $f\in\mathbb{H}$,
        for any sequence $\mathcal{I}_T$,
        the mistake bound of Ahpatron satisfies
        \begin{align*}
            &\vert M_T\vert \leq L_T(f)+\\
            &\max\left\{\frac{12U}{\sqrt{B}}L_T(f)+\overline{\Delta},\frac{0.9U}{\sqrt{B}}L_T(f)+
            \frac{\sqrt{B}}{2U}\Vert f\Vert^2_{\mathcal{H}}+\underline{\Delta}\right\},
        \end{align*}
        where
        $$
            \overline{\Delta}=\frac{\frac{3U}{\sqrt{B}}-\varepsilon}{1-\frac{3U}{\sqrt{B}}}\vert N_T\vert,
            \quad
            \underline{\Delta}=\frac{\frac{U}{\sqrt{2B}}-\varepsilon}{1-\frac{U}{\sqrt{2B}}}\vert N_T\vert.
        $$
    \end{theorem}

    Note that $\overline{\Delta}\leq 0$ and $\underline{\Delta}\leq 0$.
    Omitting the constant factors,
    we obtain a mistake bound as follows.
    $$L_T(f)+
        O\left(\frac{U}{\sqrt{B}}L_T(f)+\frac{\sqrt{B}}{U}\Vert f\Vert^2_{\mathcal{H}}+
        \max\left\{\overline{\Delta},\underline{\Delta}\right\}\right).
    $$
    Since $B\ll T$, we can obtain a mistake bound of
    $$
    L_T(f)+O\left(\frac{U}{\sqrt{B}}L_T(f)\right),
    $$
    which explicitly gives the trade-off between mistake bound and budget.

\subsection{Comparison with Previous Results}

    In the following, we assume $f\in\mathbb{H}$.
    The BOGD algorithm \cite{Zhao2012Fast} enjoys an expected regret bound of
    $O\left(\frac{UT}{\sqrt{B}}+U\sqrt{T}\right)$.
    Note that $\mathbb{I}_{\hat{y}_t\neq y_t}\leq \ell_{\mathrm{Hinge}}(f_t({\bf x}_t),y_t)$,
    we have
    \begin{align*}
    \mathbb{E}\left[\vert M_T\vert\right]=L_T(f)+
        O\left(\frac{UT}{\sqrt{B}}+U\sqrt{T}\right).
    \end{align*}
    Ahpatron significantly improves the mistake bound of BOGD, since $L_T(f)=O(T)$.
    Besides,
    our mistake bound is deterministic.

    The mistakes of POMDR \cite{Li2023Improved} satisfy
    $$
        \vert M_T\vert=L_T(f)+
        O\left(\frac{U\sqrt{\mathcal{A}_TT}}{\sqrt{B}}+\sqrt{\mathcal{A}_T}\right),
    $$
    where
    $
        \mathcal{A}_T=\sum^T_{t=1}\kappa({\bf x}_t,{\bf x}_t)
    -\frac{1}{T}{\bf Y}^\top_T{\bf K}_T{\bf Y}_T
    $
    is called ``kernel alignment'',
    ${\bf Y}_T=(y_1,\ldots,y_T)^\top$ and ${\bf K}_T$ is the kernel matrix on $\mathcal{I}_T$.
    \begin{theorem}
    \label{eq:AAAI2024:kernel_alignment:small_loss}
        Assume for any ${\bf x}\in\mathcal{X},
        \kappa\left({\bf x},{\bf x}\right)=1$.
        Let $U\geq 1$.
        Then $\mathcal{A}_T\geq \min_{f\in\mathbb{H}}L_T(f)$.
    \end{theorem}
    Theorem \ref{eq:AAAI2024:kernel_alignment:small_loss} implies $\min_{f\in \mathbb{H}}L_T(f) =O(\sqrt{\mathcal{A}_TT})$,
    and
    Ahpatron improves the mistake bound of POMDR.

    Let $U=\frac{\sqrt{B+1}}{4\sqrt{\ln(B+1)}}$.
    Then the mistake bound of Forgetron \cite{Dekel2005The} satisfies
    $$
    \vert M_T\vert\leq 4L_T(f)+2\Vert f\Vert^2_{\mathcal{H}}.
    $$
    Let $U=\frac{\sqrt{B}}{4}$ in Theorem \ref{thm:AAAI24:mistake_bound_of_H_Forgetron}.
    The mistake bound of Ahpatron satisfies
    \begin{equation}
    \label{eq:ECML2023:mistake_bound:H-Forgetron:corollary_1}
    \begin{aligned}
        &\vert M_T\vert   \leq\\
        &\max\left\{4L_T(f)+\overline{\Delta},1.3L_T(f)+
            2\Vert f\Vert^2_{\mathcal{H}}+\underline{\Delta}\right\}.
    \end{aligned}
    \end{equation}
    We can see that Ahpatron improves the mistake bound of Forgetron.
    Note that $U=\frac{\sqrt{B}}{4}
    >\frac{\sqrt{B+1}}{4\sqrt{\ln(B+1)}}$,
    where we remove the $\ln^{-\frac{1}{2}}{(B+1)}$ factor.
    Thus we solve the open problem posed by \citet{Dekel2005The}.

\subsection{A Refined Result}

    Now we explain why
    Ahpatron may obtain smaller mistake bounds via removing a half of examples.
    Let $\mathcal{J}=\left\{t\in M'_T\cup N_T: \vert S_t\vert=B\right\}$.
    At any round $t\in \mathcal{J}$, we remove a half of examples.
    We can prove that the mistake bound of Ahpatron depends on the following term
    \begin{equation}
    \label{eq:ICML2022:definition_xi}
        \frac{U+\lambda}{\lambda}\sum_{t\in \mathcal{J}}\left\Vert f_t-\bar{f}_t\right\Vert_{\mathcal{H}}
        \leq
        \frac{U+\lambda}{\lambda}\cdot\zeta U \cdot\vert \mathcal{J}\vert,
    \end{equation}
    where $\zeta\in(0,2]$.
    In the worst case,
    i.e., $\left\Vert f_t-\bar{f}_t\right\Vert_{\mathcal{H}}= 2U$,
    we have $\zeta=2$.
    It is natural that if $\bar{f}_t$ is close to $f_t$ for most of $t\in\mathcal{J}$,
    then $\zeta \ll 2$ is possible.
    In this case,
    Ahpatron enjoys smaller mistake bounds.

    \begin{theorem}[Algorithm-dependent Mistake Bound]
    \label{thm:AAAI24:refined_mistake_bound_of_H_Forgetron}
    Assume the inequality \eqref{eq:ICML2022:definition_xi} holds with $0<\zeta\leq 2$.
    Let $c_t=\frac{\Vert f_t\Vert_{\mathcal{H}}}{U}$,
    $B\geq 16$,
    $\lambda=\frac{U}{2\sqrt{B}}$.
        For any $\gamma\in(0,1)$,
        let $U\leq\frac{(1-\gamma)\sqrt{B}}{\frac{1}{4}+\frac{9}{2}\zeta}$
        and $\left(\frac{1}{4}+\frac{9\zeta}{2}\right)\frac{U}{\sqrt{B}}<\varepsilon<1$.
        Then the mistake bound of Ahpatron satisfies
        \begin{align*}
            \vert M_T\vert \leq L_T(f)+
            &\frac{1}{\gamma}\left(\frac{1}{4}+\frac{9\zeta}{2}\right)\frac{U}{\sqrt{B}}L_T(f)+\\
            &\frac{1-2\zeta}{\gamma}\frac{\Vert f\Vert^2_{\mathcal{H}}\sqrt{B}}{U}+\Delta_T(\zeta),
        \end{align*}
        where
        $$
            \Delta_T(\zeta)=\frac{\vert N_T\vert}{1-\left(\frac{1}{4}+\frac{9\zeta}{2}\right)\frac{U}{\sqrt{B}}}
            \cdot\left(\left(\frac{1}{4}+\frac{9\zeta}{2}\right)\frac{U}{\sqrt{B}}-\varepsilon\right).
        $$
    \end{theorem}

    Theorem \ref{thm:AAAI24:refined_mistake_bound_of_H_Forgetron} clearly shows that
    the mistake bound of Ahpatron depends on the value of $\zeta$,
    i.e., the distance between $\bar{f}_t$ and $f_t$.
    If $\zeta\ll 2$,
    then Ahpatron can use a larger value of $U$ and enjoy a smaller mistake bound.
    For instance,
    assume $\zeta\leq \frac{1}{16}$,
    and let $\gamma=\frac{2}{3}$ and
    $U=\frac{(1-\gamma)\sqrt{B}}{\frac{1}{4}+\frac{9}{2}\zeta}\geq\frac{\sqrt{B}}{1.6}$,
    we obtain $\frac{1}{3}<\varepsilon<1$, and
    the mistake bound of Ahpatron satisfies
    \begin{equation*}
        \vert M_T\vert\leq \frac{3}{2} L_T(f)+2.1\Vert f\Vert^2_{\mathcal{H}}+\Delta_T(\zeta),
    \end{equation*}
    which improves the result in \eqref{eq:ECML2023:mistake_bound:H-Forgetron:corollary_1},
    including:
    (i) The value of $U$ is larger, i.e., $U=\frac{\sqrt{B}}{1.6}>\frac{\sqrt{B}}{4}$.
    The larger the value of $U$ is, the larger $\mathbb{H}$ will be.
    (ii) The mistake bound is smaller in the case of $L_T(f)>\Vert f\Vert^2_{\mathcal{H}}$.
    \begin{remark}
        Theorem \ref{thm:AAAI24:refined_mistake_bound_of_H_Forgetron}
        also gives a criterion to select $\bar{f}_t$.
        To be specific,
        the optimal $\bar{f}_t$ must be
        $$
        \mathop{\arg\min}_{f\in \mathcal{H}_{t,2},
        S_{t,2}\subseteq S_t,\vert S_{t,2}\vert=\frac{B}{2}}\left\Vert f-f_t\right\Vert_{\mathcal{H}}.
        $$
        However,
        obtaining the optimal solution is computationally infeasible.
        Our algorithm constructs an approximation solution using a heuristic method.
    \end{remark}

\section{Experiments}

    We aim to verify Ahpatron performs better than the state-of-the-art algorithms on the same budget
    or a smaller budget.

    We adopt the Gaussian kernel $\kappa(\mathbf{u},\mathbf{v})
    =\exp(-\frac{\Vert \mathbf{u}-\mathbf{v}\Vert^2_2}{2\sigma^2})$,
    where $\sigma$ is the width.
    We download six binary classification datasets
    from UCI machine learning repository
    \footnote{\url{http://archive.ics.uci.edu/ml/datasets.php}}
    and LIBSVM website
    \footnote{\url{https://www.csie.ntu.edu.tw/~cjlin/libsvmtools/datasets/binary.html}},
    as shown in Table \ref{tab:JMLR2022:Datasets}.
    We uniformly select $50000$ instances from the original \textit{SUSY} dataset.
    All algorithms are implemented in R on a Windows machine with
    2.8 GHz Core(TM) i7-1165G7 CPU
    \footnote{Codes and datasets: \url{https://github.com/alg4ml/Ahpatron.git}}.
    We execute each experiment 10 times with random permutation of all datasets.

    \begin{table}[!t]
    \footnotesize
      \centering
      \setlength{\tabcolsep}{1.3mm}
      \begin{tabular}{l|rrr}
        \toprule
        Datasets    &\#sample &\# Feature &Classes        \\
        \hline
        phishing    & 11,055 & 68  & 2\\
        a9a         & 48,842 & 123 & 2\\
        w8a         & 49,749 & 300 & 2\\
        SUSY        & 50,000 &  18 & 2\\
        ijcnn1      & 141,691 & 22 & 2\\
        cod-rna     & 271,617 & 8  & 2\\
        \bottomrule
      \end{tabular}
      \caption{Datasets used in the experiments}
      \label{tab:JMLR2022:Datasets}
    \end{table}



\begin{table*}[!t]
\footnotesize
      \centering
      \setlength{\tabcolsep}{1.8mm}
      \begin{tabular}{l|rrr|rrr|rrr}

        \toprule
        \multirow{2}{*}{Algorithm}&\multicolumn{3}{c|}{phishing, $\sigma=1$}
                                  &\multicolumn{3}{c|}{SUSY, $\sigma=1$}
                                  &\multicolumn{3}{c}{ijcnn1, $\sigma=1$}\\
        \cline{2-10}&AMR (\%) &$B|D$       &Time (s)  &AMR (\%) &$B|D$       &Time (s)
        &AMR (\%) &$B|D$       &Time (s)       \\
        \hline
        Projectron++         & 7.42 $\pm$ 0.15  & \textbf{571}   & 3.04
                             & 22.05 $\pm$ 0.07 & \textbf{4898}  & 1330.83
                             & 3.01 $\pm$ 0.04  & \textbf{1577}  & 58.58\\
        Projectron           & 9.52 $\pm$ 0.16  & 422   & 1.71
                             & 29.10 $\pm$ 0.17 & 2111 & 108.87
                             & 4.14 $\pm$ 0.04   & 1136  & 28.22\\
        FOGD                 & 7.48 $\pm$ 0.16  & 2000  & 3.30
                             & 26.70 $\pm$ 0.60 & 2000 & 10.04
                             & 3.74 $\pm$ 0.23   & 2000  & 19.12\\
        BOGD++               & 10.30 $\pm$ 0.20 & 400   & 1.01
                             & 28.30 $\pm$ 0.12 & 400  & 3.09
                             & 9.21 $\pm$ 0.02   & 600   & 10.55\\
        NOGD                 & 8.12 $\pm$ 0.23  & 400   & 3.02
                             & 25.49 $\pm$ 0.47 & 400  & 3.92
                             & 4.47 $\pm$ 0.06   & 600   & 20.71\\
        POMDR                & \textbf{7.36}  $\pm$ \textbf{0.16}  & 400  & \textbf{1.29}
                             & 25.36 $\pm$ 0.25 & 400 & 2.79
                             & 5.76  $\pm$ 0.09  & 600   & 9.33\\
        Ahpatron          & \textbf{7.27} $\pm$ \textbf{0.13}  & 400   & \textbf{1.01}
                             & \textbf{23.66} $\pm$ \textbf{0.17} & 400 & \textbf{3.62}
                             & \textbf{3.72}  $\pm$ \textbf{0.03}  & 600   & \textbf{8.51}\\
        \midrule
        \multirow{2}{*}{Algorithm}&\multicolumn{3}{c|}{cod-rna, $\sigma=1$}
                                  &\multicolumn{3}{c|}{w8a, $\sigma=2$}
                                  &\multicolumn{3}{c}{a9a, $\sigma=2$}\\
        \cline{2-10}&AMR (\%) &$B|D$       &Time (s)  &AMR (\%) &$B|D$       &Time (s)
        &AMR (\%) &$B|D$       &Time (s) \\
        \hline
        Projectron++         & 10.13 $\pm$ 0.03  & \textbf{5100}  & 2786.98
                             & 2.03  $\pm$ 0.03  & \textbf{1279}  & 94.37
                             & \textbf{16.45} $\pm$ \textbf{0.07}  & 152   & 4.22\\
        Projectron           & 14.43 $\pm$ 0.03  & 2686  & 453.49 & 2.90  $\pm$ 0.06 & 306  & 14.15
                             & 21.13 $\pm$ 0.17  & 135   & 3.46\\
        FOGD                 & 12.72 $\pm$ 1.05  & 2000  & 47.41  & 4.18 $\pm$ 0.23  & 300  & 6.28
                             & 16.69 $\pm$ 0.11  & 300   & 3.71  \\
        BOGD++               & 14.01 $\pm$ 0.04  & 600   & 12.78  & 2.76 $\pm$ 0.08  & 300  & 14.25
                             & 19.91 $\pm$ 0.12  & 150   & 3.34 \\
        NOGD                 & 15.85 $\pm$ 0.72  & 600   & 30.61  & 2.69 $\pm$ 0.04  & 300  & 15.04
                             & 16.56 $\pm$ 0.15  & 150   & 3.56  \\
        POMDR                & \textbf{11.76} $\pm$ \textbf{0.11} & 600  & \textbf{11.58}
                             & 2.59 $\pm$ 0.11   & 300   & 15.00
                             & 17.09 $\pm$ 0.15  & 150   & 4.33 \\
        Ahpatron             & 12.33 $\pm$ 0.08  & 600   & 14.85
                             & \textbf{2.23} $\pm$ \textbf{0.06}  & 300   & \textbf{11.68}
                             & \textbf{16.42} $\pm$ \textbf{0.06}  & 150   & 3.14 \\
        \bottomrule
      \end{tabular}
      \caption{Comparison with the state-of-the-art algorithms}
      \label{tab:ML2021:memory_constraint:classification_different_budget}
    \end{table*}

    The state-of-the-art algorithms we compare with are listed as follows.
    \begin{itemize}
      \item Projectron and Projectron++ \cite{Orabona2008The}\\
            The two algorithms do not remove examples,
            but use the approximate linear dependence condition \cite{Engel2004The} to add examples.
            The two algorithms may maintain a very large budget
            and thus suffer high memory costs and running time.
      \item BOGD++ \cite{Zhao2012Fast}, POMDR \cite{Li2023Improved}\\
            When the size of the active set reaches the budget,
            BOGD++ randomly removes one example,
            while POMDR removes a half of the examples.
      \item NOGD \cite{Lu2016Large}\\
            NOGD first selects $B$ examples and constructs the corresponding kernel matrix ${\bf K}_{B}$.
            Then it constructs a new representation in $\mathbb{R}^k$ for each instance
            by the best rank-$k$ approximation of ${\bf K}_{B}$,
            where $k\leq B$.
            The space complexity of NOGD is in $O\left(\max\left\{B^2,Bd\right\}\right)$.
      \item FOGD \cite{Lu2016Large}\\
            FOGD uses random features \cite{Rahimi2007Random} to approximate kernel functions,
            and runs online gradient descent in $\mathbb{R}^D$.
            $D$ is the number of random features.
            The computational complexity of FOGD is in $O(Dd)$.
    \end{itemize}
    We do not compare with Forgetron and RBP,
    since their performances are much worse than FOGD and NOGD.
    For BOGD++, NOGD, and FOGD,
    we choose the stepsize of gradient descent from $\left\{\frac{10^{[-3:1:3]}}{\sqrt{T}}\right\}$.
    The other parameters of BOGD++ and NOGD follow the original paper.
    All parameters of POMDR also follow the original paper.
    For Projectron and Projectron++,
    there is a parameter $0<\eta<1$
    balancing the memory costs and prediction performance.
    We choose $\eta\in\{0.1,0.9\}$.
    The smaller $\eta$ is, the larger the size of the active set is
    and the better the prediction performance is.
    For Ahpatron,
    we set the parameters following
    Theorem \ref{thm:AAAI24:refined_mistake_bound_of_H_Forgetron},
    that is, $\eta=0.0005,\lambda=\frac{U}{\sqrt{4B}}$ and $U=\frac{\sqrt{B}}{2}$.
    We choose the best $\varepsilon\in\{0.5,0.6,0.7,0.8,0.9\}$ in hindsight,
    and set $\sigma=1$ for all datasets.
    If the per-round running time of Projectron++ is larger than $1$ hour,
    then we set $\sigma=2$.

    Table \ref{tab:ML2021:memory_constraint:classification_different_budget}
    shows the results.
    We record the average mistake rates (AMR)
    and $\mathrm{AMR}:=\frac{\vert M_T\vert}{T}$.
    As a whole,
    Ahpatron performs best on most of datasets on a small budget.
    Although there are some datasests on which
    Projectron++ enjoys a lower AMR than Ahpatron,
    the budget maintained by Projectron++ is very large,
    such as on the \textit{SUSY}, \textit{ijcnn1}, \textit{cod-rna} and \textit{w8a} datasets.
    The main reason is that Projectron and Projectron++ only add examples.
    In practice,
    it must be careful to choose Projectron++ and Projectron.
    In contrast, Ahpatron can precisely control the budget
    and show a comparable performance on all datasets.
    Specially,
    Ahpatron even performs better than Projection++
    on the \textit{phishing} datasets using a smaller budget.
        In the supplementary material,
    we further compare with Projectron++ on
    the \textit{SUSY}, \textit{ijcnn1}, \textit{cod-rna} and \textit{w8a} dataset.
    We aim to show that
    Ahpatron performs better than Projectron++
    on the same or a smaller budget.

    Ahpatron performs better than BOGD++ on all datsets.
    BOGD++ approximates gradient descent and removes one example,
    while Ahpatron approximates AVP and removes a half of examples.
    The results prove that AVP is better than gradient descent
    and our budget maintaining approach is also better.
    Ahpatron also performs better than POMDR on most of datasets,
    since it adopts a different strategy to remove the examples
    and projection scheme to keep the information of the removed examples.
    Ahpatron also performs better than NOGD and FOGD on all datasets.

    We also report the value of $\left\vert N_T\right\vert$ in Ahpatron.
    Table \ref{tab:JMLR2022:N_T} shows the results.
    It is obvious that $\left\vert N_T\right\vert \gg 1$ on all datasets.
    Thus the negative terms in Theorem \ref{thm:AAAI24:mistake_bound_of_H_Forgetron}
    and Theorem \ref{thm:AAAI24:refined_mistake_bound_of_H_Forgetron}
    can significantly reduce the mistake bound.

    \begin{table}[!t]
    \footnotesize
      \centering
      \setlength{\tabcolsep}{1.5mm}
      \begin{tabular}{l|rrrrrr}
        \toprule
        &phishing &SUSY &cod-rna&w8a&ijcnn1 &a9a         \\
        \hline
        $\vert N_T\vert$& 1429 & 23045 & 129497 &1525&5204 & 9234   \\
        \bottomrule
      \end{tabular}
      \caption{The value of $\vert N_T\vert$ in Ahpatron}
      \label{tab:JMLR2022:N_T}
    \end{table}

\section{Conclusion}
In this paper,
we have proposed a new budgeted online kernel learning model,
called Ahpatron,
which has a tighter mistake bound,
and resolved the open problem posed by \citet{Dekel2005The}.
The key of Ahpatron is the half removing and half projecting budget mechanism,
which keeps as more information of the removed examples as possible.
We also have proved that the active updating rule can improve the mistake bound.

The mistake bound of Ahpatron explicitly gives the trade-off
between the mistake bound and the budget,
which is important for online learning,
and is left for future research.

\section*{Acknowledgements}
    This work was supported in part by the National Natural Science Foundation of China (No.62076181).
    We thank all anonymous reviewers for their valuable comments and suggestions.

\bibliography{aaai24}

\begin{thebibliography}{25}
\providecommand{\natexlab}[1]{#1}

\bibitem[{Auer, Cesa{-}Bianchi, and Gentile(2002)}]{Auer2002Adaptive}
Auer, P.; Cesa{-}Bianchi, N.; and Gentile, C. 2002.
\newblock Adaptive and Self-Confident On-Line Learning Algorithms.
\newblock \emph{Journal of Computer and System Sciences}, 64(1): 48--75.

\bibitem[{Cesa{-}Bianchi and Gentile(2006)}]{Cesa-Bianchi2006Tracking}
Cesa{-}Bianchi, N.; and Gentile, C. 2006.
\newblock Tracking the best hyperplane with a simple budget perceptron.
\newblock In \emph{Proceedings of the 19th Annual Conference on Learning
  Theory}, 483--498.

\bibitem[{Charikar, Chen, and Farach{-}Colton(2002)}]{Charikar2002Finding}
Charikar, M.; Chen, K.~C.; and Farach{-}Colton, M. 2002.
\newblock Finding frequent items in data streams.
\newblock In \emph{Proceedings of the 29th International Colloquium on
  Automata, Languages and Programming}, 693--703.

\bibitem[{Cheng et~al.(2006)Cheng, Vishwanathan, Schuurmans, Wang, and
  Caelli}]{Cheng2006implicit}
Cheng, L.; Vishwanathan, S. V.~N.; Schuurmans, D.; Wang, S.; and Caelli, T.
  2006.
\newblock Implicit online learning with kernels.
\newblock \emph{Advances in Neural Information Processing Systems}, 19:
  249--256.

\bibitem[{Crammer, Kandola, and Singer(2003)}]{Crammer2003Online}
Crammer, K.; Kandola, J.~S.; and Singer, Y. 2003.
\newblock Online classification on a budget.
\newblock \emph{Advances in Neural Information Processing Systems}, 16:
  225--232.

\bibitem[{Dekel, Shalev{-}Shwartz, and Singer(2005)}]{Dekel2005The}
Dekel, O.; Shalev{-}Shwartz, S.; and Singer, Y. 2005.
\newblock The {F}orgetron: {A} kernel-based perceptron on a fixed budget.
\newblock \emph{Advances in Neural Information Processing Systems}, 18:
  259--266.

\bibitem[{Dekel, Shalev{-}Shwartz, and Singer(2008)}]{Dekel2008The}
Dekel, O.; Shalev{-}Shwartz, S.; and Singer, Y. 2008.
\newblock The {F}orgetron: {A} kernel-based perceptron on a budget.
\newblock \emph{SIAM Journal on Computing}, 37(5): 1342--1372.

\bibitem[{Engel, Mannor, and Meir(2004)}]{Engel2004The}
Engel, Y.; Mannor, S.; and Meir, R. 2004.
\newblock The kernel recursive least-squares algorithm.
\newblock \emph{IEEE Transactions on Signal Processing}, 52(8): 2275--2285.

\bibitem[{Gentile(2001)}]{Gentile2001A}
Gentile, C. 2001.
\newblock A new approximate maximal margin classification algorithm.
\newblock \emph{Journal of Machine Learning Research}, 2: 213--242.

\bibitem[{He and Kwok(2014)}]{He2014Simple}
He, W.; and Kwok, J.~T. 2014.
\newblock Simple randomized algorithms for online learning with kernels.
\newblock \emph{Neural Networks}, 60: 17--24.

\bibitem[{Kivinen, Smola, and Williamson(2001)}]{Kivinen2001Online}
Kivinen, J.; Smola, A.~J.; and Williamson, R.~C. 2001.
\newblock Online learning with kernels.
\newblock \emph{Advances in Neural Information Processing Systems}, 14:
  785--792.

\bibitem[{Koppel et~al.(2019)Koppel, Warnell, Stump, and
  Ribeiro}]{Koppel2019Parsimonious}
Koppel, A.; Warnell, G.; Stump, E.; and Ribeiro, A. 2019.
\newblock Parsimonious online learning with kernels via sparse projections in
  function space.
\newblock \emph{Journal of Machine Learning Research}, 20(3): 1--44.

\bibitem[{Li and Liao(2023)}]{Li2023Improved}
Li, J.; and Liao, S. 2023.
\newblock Improved kernel alignment regret bound for online kernel kearning.
\newblock In \emph{Proceedings of the Thirty-Seventh {AAAI} Conference on
  Artificial Intelligence}, 8597--8604.

\bibitem[{Lu et~al.(2016)Lu, Hoi, Wang, Zhao, and Liu}]{Lu2016Large}
Lu, J.; Hoi, S. C.~H.; Wang, J.; Zhao, P.; and Liu, Z. 2016.
\newblock Large scale online kernel learning.
\newblock \emph{Journal of Machine Learning Research}, 17(47): 1--43.

\bibitem[{Orabona, Keshet, and Caputo(2008)}]{Orabona2008The}
Orabona, F.; Keshet, J.; and Caputo, B. 2008.
\newblock The {P}rojectron: {A} bounded kernel-based Perceptron.
\newblock In \emph{Proceedings of the Twenty-Fifth International Conference on
  Machine Learning}, 720--727.

\bibitem[{Rahimi and Recht(2007)}]{Rahimi2007Random}
Rahimi, A.; and Recht, B. 2007.
\newblock Random features for large-scale kernel machines.
\newblock \emph{Advances in Neural Information Processing Systems}, 20:
  1177--1184.

\bibitem[{Rosenblatt(1958)}]{Frank1958The}
Rosenblatt, F. 1958.
\newblock The {P}erceptron: {A} probabilistic model for information storage and
  organization in the brain.
\newblock \emph{Psychological Review}, 65: 386--408.

\bibitem[{Shalev-Shwartz(2007)}]{Shai2007Online}
Shalev-Shwartz, S. 2007.
\newblock \emph{Online learning: {T}heory, algorithms, and applications}.
\newblock Ph.D. thesis, The Hebrew University of Jerusalem.

\bibitem[{Vapnik(1998)}]{vapnik1998statistical}
Vapnik, V.~N. 1998.
\newblock \emph{Statistical learning theory}, volume~1.
\newblock New York: Wiley \& Sons.

\bibitem[{Wang et~al.(2013)Wang, Hoi, Zhao, Zhuang, and Liu}]{Wang2013Large}
Wang, J.; Hoi, S. C.~H.; Zhao, P.; Zhuang, J.; and Liu, Z. 2013.
\newblock Large scale online kernel classification.
\newblock In \emph{Proceedings of the 23rd International Joint Conference on
  Artificial Intelligence}, 1750--1756.

\bibitem[{Wang, Crammer, and Vucetic(2012)}]{Wang2012Breakingecond}
Wang, Z.; Crammer, K.; and Vucetic, S. 2012.
\newblock Breaking the curse of kernelization: budgeted stochastic gradient
  descent for large-scale {SVM} training.
\newblock \emph{Journal of Machine Learning Research}, 13(1): 3103--3131.

\bibitem[{Weston, Bordes, and Bottou(2005)}]{Weston2005Online}
Weston, J.; Bordes, A.; and Bottou, L. 2005.
\newblock Online (and offline) on an even tighter budget.
\newblock In \emph{Proceedings of the Tenth International Workshop on
  Artificial Intelligence and Statistics}, 413--420.

\bibitem[{Williams and Seeger(2001)}]{Williams2001}
Williams, C. K.~I.; and Seeger, M. 2001.
\newblock Using the {N}ystr\"{o}m method to speed up kernel machines.
\newblock \emph{Advances in Neural Information Processing Systems}, 13:
  682--688.

\bibitem[{Zhang and Liao(2019)}]{Zhang2019Incremental}
Zhang, X.; and Liao, S. 2019.
\newblock Incremental randomized sketching for online kernel learning.
\newblock In \emph{Proceedings of the 36th International Conference on Machine
  Learning}, 7394--7403.

\bibitem[{Zhao et~al.(2012)Zhao, Wang, Wu, Jin, and Hoi}]{Zhao2012Fast}
Zhao, P.; Wang, J.; Wu, P.; Jin, R.; and Hoi, S. C.~H. 2012.
\newblock Fast bounded online gradient descent algorithms for scalable
  kernel-based online learning.
\newblock In \emph{Proceedings of the 29th International Conference on Machine
  Learning}, 1075--1082.

\end{thebibliography}


\section{Notations}

    Table \ref{tab:AAAI24:Notations} gives the notations used in the manual and the following proof.

    \begin{table}[!ht]
    \footnotesize
      \centering
      \setlength{\tabcolsep}{1.3mm}
      \caption{Notation table}
      \label{tab:AAAI24:Notations}
      \begin{tabular}{ll}
        \toprule
        Notation    &Definition \\
        \hline
        $M_T$    & $\left\{t\in[T]: \hat{y}_t\neq y_t\right\}$ \\
        $M'_T$   & $\left\{t\in[T]: y_t f_t\left({\bf x}_t \right)\leq 0\right\}$, $M'_T\supseteq M_T$\\
        $N_T$    & $\left\{t\in[T]: 0<y_tf_t\left({\bf x}_t \right) < 1-\varepsilon_t\right\}$\\
        $\ell_{\mathrm{Hinge}}(u,y)$ & $\max\{0,1-uy\}$, the Hinge loss function\\
        $L_T(f)$ & $\sum^T_{t=1}\ell_{\mathrm{Hinge}}\left(f\left({\bf x}_t\right),y_t\right)$\\
        $\nabla_f \ell_{\mathrm{Hinge}}\left(f\left({\bf x}_t \right),y_t\right)$
        &the gradient of $\ell_{\mathrm{Hinge}}\left(f\left({\bf x}_t\right),y_t\right)$ w.r.t. $f$\\
        $\nabla_t$ & $\nabla_{f_t} \ell_{\mathrm{Hinge}}\left(f_t\left({\bf x}_t\right),y_t\right)$\\
        $\mathbb{H}$ & $\left\{f\in\mathcal{H}:\Vert f\Vert_{\mathcal{H}}\leq U\right\}$\\
        $\mathrm{Proj}_{\mathbb{H}}\left(g\right)$  & the projection of $g$ on $\mathbb{H}$\\
        $\mathcal{A}_T$ & the kernel alignment complexity\\
        $S_t$           & the active set\\
        \bottomrule
      \end{tabular}
    \end{table}

\section{Proof of Theorem \ref{thm:ECML2023:Perceptron++}}

    \begin{proof}
        The updating rule of AVP is as follows.
        \begin{align*}
            \tilde{f}_{t+1}=&f_t+\lambda_t\cdot y_t\kappa\left({\bf x}_t,\cdot\right)\cdot
            \mathbb{I}_{y_t\cdot f_t(\mathbf{x}_t)<1-\varepsilon_t},\\
            f_{t+1}=&\mathrm{Proj}_{\mathbb{H}}\left(\tilde{f}_{t+1}\right).
        \end{align*}
        Note that $\varepsilon_t>0$.
        At any round $t$,
        if $y_tf_t({\bf x}_t)<1-\varepsilon_t$,
        then it must be $\ell_{\mathrm{Hinge}}\left(f_t\left({\bf x}_t\right),y_t\right)>0$.

        Using the convexity of the hinge loss function,
        we have
        \begin{align}
            &\ell_{\mathrm{Hinge}}\left(f_t\left({\bf x}_t\right),y_t\right)-
             \ell_{\mathrm{Hinge}}\left(f\left({\bf x}_t\right),y_t\right)\nonumber\\
            \leq
            & \left\langle \nabla_t,f_t-f \right\rangle \nonumber\\
            =&\frac{1}{\lambda_t}\left\langle f_t- \tilde{f}_{t+1},f_t-f\right\rangle\nonumber\\
            =&\frac{1}{2\lambda_t}\left[
                                        \left\Vert f_t - f \right\Vert^2_{\mathcal{H}} -
                                        \left\Vert \tilde{f}_{t+1}- f \right\Vert^2_{\mathcal{H}} +
                                        \left\Vert \tilde{f}_{t+1} - f_t\right\Vert^2_{\mathcal{H}}
                                  \right]\nonumber\\
            =&\frac{\left \Vert f_t - f\right \Vert^2_{\mathcal{H}} -
                    \left \Vert \tilde{f}_{t+1} - f \right\Vert^2_{\mathcal{H}}}
                   {2\lambda_t}
              +\frac{\lambda_t}{2} \left\Vert \nabla_t\right\Vert^2_{\mathcal{H}},
               \label{eq:Supp_ECML2023:analysis_instantaneous_regret}
        \end{align}
        where $\nabla_t = -y_t \kappa\left({\bf x}_t,\cdot\right)$.

        Recalling the definition of $M'_T$ and $N_T$ in Table \ref{tab:AAAI24:Notations},
        and summing over $t\in M'_T\cup N_T$ gives
        \begin{align*}
            &\sum_{t\in M'_T\cup N_T}\left[\ell_{\mathrm{Hinge}}\left(f_t\left({\bf x}_t\right),y_t\right) -
                                           \ell_{\mathrm{Hinge}}\left(f\left({\bf x}_t\right),y_t\right)\right]\\
            \leq
            & \sum_{t\in M'_T\cup N_T}\frac{\left\Vert f_t - f \right\Vert^2_{\mathcal{H}} -
                                            \left\Vert f_{t+1} - f \right\Vert^2_{\mathcal{H}}}{2\lambda_t} +
              \sum_{t\in M'_T\cup N_T}\frac{\lambda_t}{2},
        \end{align*}
        where we use the assumption $\kappa(\cdot,\cdot)\leq 1$ and
        the property of the projection.

        Denote by
        $$
        \delta_t = \left\Vert f_t-f \right\Vert^2_{\mathcal{H}}
            - \left\Vert f_{t+1}-f\right\Vert^2_{\mathcal{H}}.
        $$

        Let
        $$
        1> \varepsilon_t> \dfrac{\lambda_{t}}{2},t\geq 1, \lambda_t<2.
        $$

        For all $t\in M'_T$,
        $$
        \ell_{\mathrm{Hinge}}\left(f_t\left({\bf x}_t\right),y_t\right) \geq \mathbb{I}_{\hat{y}_t\neq y_t}.
        $$
        We can derive the following inequality,
        \begin{align}
             &\sum_{t\in M'_T\cup N_T} \ell_{\mathrm{Hinge}}\left(f_t\left({\bf x}_t\right),y_t\right) -
                                      \left\vert M'_T \right\vert\nonumber\\
        \geq &\sum_{t\in N_T} \ell_{\mathrm{Hinge}}\left(f_t\left({\bf x}_t\right),y_t\right)\nonumber\\
        =    &\sum_{t\in N_T}\left(1-y_t\cdot f_t\left({\bf x}_t\right)\right)
              \geq \sum_{t\in N_T}\varepsilon_t.
            \label{eq:Supp_ECML2023:gap:losses_and_mistakes}
        \end{align}

        Then we obtain
        \begin{align*}
                &\left\vert M'_T\right\vert - L_T(f)\\
        \leq    &\sum_{t\in M'_T \cup N_T}\left[\ell_{\mathrm{Hinge}}\left(f_t\left({\bf x}_t\right),y_t\right) -
                                                \ell_{\mathrm{Hinge}}\left(f\left({\bf x}_t\right),y_t\right)\right]-\\
                &\sum_{t\in N_T}\varepsilon_t\\
        \leq    &\sum_{t\in M'_T\cup N_T} \frac{\delta_t}{2\lambda_t}+
                 \sum_{t\in M'_T}\frac{\lambda_t}{2}+
                 \sum_{t\in N_T}\left(\frac{\lambda_t}{2}-\varepsilon_t\right),
        \end{align*}
        which concludes the proof.
    \end{proof}

\section{Proof of Corollary \ref{coro:ECML2023:perceptron++:constant_learnining_rate}}

    \begin{proof}
        A key observation is as follows.
        \begin{quote}
          For any $t \notin M'_T\cup N_T$,
          \begin{equation}
            \label{eq:Supp_ECML2023:Porgetron++:aux_lemma1}
            f_t=f_{t'},
        \end{equation}
        where $t'=\min_{\tau}\left\{\tau>t:\tau\in M'_T\cup N_T\right\}$.
        \end{quote}

        Let $M'_T\cup N_T=\{t_1,t_2,\ldots,t_{K}\}$,
        where $K=\vert M'_T\vert+\vert N_T\vert$.

        If $t_{i-1}+1\in M'_T\cup N_T$,
        then it must be $t_{i-1}+1=t_i$ and $f_{t_{i-1}+1} = f_{t_i}$.
        The reason is that $t_{i-1}<t_i$.
        If $t_{i-1}+1\notin M'_t\cup N_T$,
        then according to \eqref{eq:Supp_ECML2023:Porgetron++:aux_lemma1},
        we obtain $f_{t_{i-1}+1}=f_{t_i}$.

        For any $t\in[T]$, let $\lambda_t=1$ and $\varepsilon_t=\varepsilon$.
        Given $f\in\mathcal{H}$, using the result in Theorem 3,
        we obtain
        \begin{align*}
                &\vert M'_T\vert -L_T(f)\\
        \leq    &\sum_{t\in M'_T\cup N_T}\frac{\delta_t}{2}+
                 \frac{1}{2}\vert M'_T\vert+
                 \left(\frac{1}{2}-\varepsilon\right) \cdot \left\vert N_T \right\vert\\
        =       &\frac{\left\Vert f_{t_1}-f \right\Vert^2_{\mathcal{H}}}{2}+
                 \frac{1}{2}\left\vert M'_T\right\vert
            +\left(\frac{1}{2}-\varepsilon\right)\cdot\vert N_T\vert.
        \end{align*}

        Note that $f_{t_1}=f_1=0$.

        Rearranging terms yields
        $$
            \left\vert M'_T \right\vert
            \leq
            2L_T(f)+\Vert f\Vert^2_{\mathcal{H}}+\left(1-2\varepsilon\right)\cdot\left\vert N_T\right\vert.
        $$
        Using the fact $\left\vert M_T\right\vert \leq \left\vert M'_T\right\vert$ concludes the proof.
    \end{proof}

\section{Proof of Corollary \ref{coro:ECML2023:ALMA++:time_learnining_rate}}

    \begin{proof}
        We first restate a technical lemma.
        \begin{lemma}[Lemma 3.5 in \citet{Auer2002Adaptive}]
             Let $l_1,l_2,\ldots,l_T$ and $\xi$ be non-negative real-number.
             Then
             $$
                \sum^T_{t=1}\frac{l_t}{\sqrt{\xi+\sum^t_{\tau}l_\tau}}\leq 2\sqrt{\xi+\sum^T_{t=1}l_t}-2\sqrt{\xi}.
             $$
        \end{lemma}

        Recalling that for any $t\geq 1$ the learning rates are defined as follows.
        $$
            \lambda_t=\frac{U}{\sqrt{U^2+\sum^t_{\tau=1}\mathbb{I}_{y_{\tau}f_{\tau}({\bf x}_\tau)\leq 0}}}.
        $$

        Denote by
        $$
        g_t=\left\Vert f_t -f\right\Vert^2_{\mathcal{H}}.
        $$

        We have $g_t\leq 4U^2$ and
        \begin{align*}
            &\sum_{t\in M'_T\cup N_T}\frac{\delta_t}{2\lambda_t}\\
            =&\frac{g_{t_1}}{2\lambda_{t_1}}
              +\sum^K_{i=2} \left[\frac{g_{t_i}}{2\lambda_{t_i}}- \frac{g_{t_{i-1}+1}}{2\lambda_{t_{i-1}}}\right]
              -\frac{g_{t_K+1}}{2\lambda_{t_K}}\\
            =&\frac{g_{t_1}}{2\lambda_{t_1}}
              +\sum^K_{i=2}g_{t_i}\left[\frac{1}{2\lambda_{t_i}}-\frac{1}{2\lambda_{t_{i-1}}}\right]
              -\frac{g_{t_K+1}}{2\lambda_{t_K}}\\
            \leq
            &\frac{2U^2}{\min\limits_{t\in M'_T\cup N_T}\lambda_t}\\
            \leq
            & 2\sqrt{U^2+\left\vert M'_T\right\vert}.
        \end{align*}

        Using the result in Theorem 3,
        we obtain
        \begin{align*}
                &\left\vert M'_T\right\vert -L_T(f)\\
        \leq    &\sum_{t\in M'_T\cup N_T}\frac{\delta_t}{2\lambda_t}
                 +\sum_{t\in M'_T}\frac{\lambda_t}{2}
                 +\sum_{t\in N_T}\left(\frac{\lambda_t}{2}-\varepsilon_t\right)\\
        \leq    &2U\sqrt{U^2+\left\vert M'_T\right\vert}
                 +U\sqrt{\left\vert M'_T \right\vert }
                 +\sum_{t\in N_T}\left(\frac{\lambda_t}{2}-\varepsilon_t\right)\\
        \leq    &2U^2+3U\sqrt{\left\vert M'_T\right\vert}
                 +\sum_{t\in N_T}\left(\frac{\lambda_t}{2}-\varepsilon_t\right),
        \end{align*}
        where the second inequality comes from the above Lemma. 

        If
        $$
            L_T(f)+2U^2
            +\sum_{t\in N_T}\left(\frac{\lambda_t}{2}-\varepsilon_t\right) \leq 0,
        $$
        then
        $$
            \left\vert M'_T \right\vert \leq 9U^2.
        $$
        Otherwise, solving for $\left\vert M'_T \right\vert$ gives
        \begin{align*}
                &\left\vert M'_T \right\vert -L_T(f)\\
        \leq    &9U^2+\sum_{t\in N_T}\left(\frac{\lambda_t}{2}-\varepsilon_t\right)+2U^2+\\
                &3U\sqrt{L_T(f)+2U^2+\sum_{t\in N_T}\left(\frac{\lambda_t}{2}-\varepsilon_t\right)}.
        \end{align*}
        Combining with the two cases,
        and using the fact $\vert M_T\vert \leq \vert M'_T\vert$ concludes the proof.
    \end{proof}

\section{Proof of Lemma \ref{lemma:ECML2023:number_of_removing}}

    \begin{proof}
        At any round $t$,
        if
        $$
        \left\vert S_t \right\vert=B \quad \textrm{and}\quad y_t\cdot f_t({\bf x}_t)< 1-\varepsilon_t,
        $$
        then we must remove $\frac{B}{2}$ examples.

        Let
        $$
        \mathcal{J}=\left\{\bar{t}_1,\bar{t}_2,\ldots,\bar{t}_{\vert \mathcal{J}\vert}\right\}
        $$
        be the set of time indexes
        such that
        $$
        \forall t\in \mathcal{J},~
        \left[\left(\left\vert S_t \right\vert = B \right) \wedge \left(y_t\cdot f_t\left({\bf x}_t\right)< 1-\varepsilon_t\right)\right].
        $$

        For simplicity, let $J=\vert \mathcal{J}\vert$.

        At any round $t$,
        if $y_t\cdot f_t({\bf x}_t) < 1-\varepsilon_t$,
        then we add the current example into $S_t$.
        Recalling that
        $$
            M'_T\cup N_T=\{t\in[T]: y_t\cdot f_t({\bf x}_t)< 1-\varepsilon_t\}.
        $$

        Next we consider two cases.

        \begin{itemize}
          \item \textbf{Case 1}: $\left\vert M'_T\right\vert + \left\vert N_T \right\vert \leq B$.\\
                No examples will be removed and $J=0$.
          \item \textbf{Case 2}: $\left\vert M'_T\right\vert + \left\vert N_T\right\vert> B$.\\
                It is obvious that $\bar{t}_1\geq B$.
                For any $j\in\{1,\ldots,J-1\}$,
                there are $\frac{B}{2}$ examples that are added into the budget
                in the time interval $\{\bar{t}_j+1,\bar{t}_j+2,\bar{t}_{j+1}\}$.
                For $1\leq t\leq \bar{t}_1$, there are $B$ examples that are added into the budget.
                For $\bar{t}_J<t\leq T$,
                there are $\frac{B}{2}$ examples that are added into the budget at most.
                Thus we can derive the following inequality
                $$
                    B+\frac{B}{2}(J-1) \leq  \left\vert M'_T \right\vert + \left\vert N_T \right\vert,
                $$
                which implies $J\leq \frac{2\left(\left\vert M'_T\right\vert+\left\vert N_T\right\vert\right)}{B}-1$.
        \end{itemize}
        Combining with the two cases concludes the proof.
    \end{proof}

\section{Proof of Theorem \ref{thm:AAAI24:mistake_bound_of_H_Forgetron}}

    \begin{proof}
        We first consider the case $J\geq 1$.

        For any $t\in M'_T\cup N_T\setminus \mathcal{J}$,
        we start with \eqref{eq:Supp_ECML2023:analysis_instantaneous_regret}.

        Summing over $t\in M'_T\cup N_T\setminus \mathcal{J}$ gives
        \begin{align*}
                &\sum_{t\in M'_T\cup N_T\setminus \mathcal{J}}\left[\ell_{\mathrm{Hinge}}\left(f_t\left({\bf x}_t\right),y_t\right)
                 -\ell_{\mathrm{Hinge}}\left(f\left({\bf x}_t\right),y_t\right)\right]\\
            \leq&\sum_{t\in M'_T\cup N_T\setminus \mathcal{J}}\frac{\left\Vert f_t-f\right\Vert^2_{\mathcal{H}}
                 -\left\Vert f_{t+1}-f\right\Vert^2_{\mathcal{H}}}{2\lambda}+\\
                &\frac{\lambda}{2}\left\vert M'_T\cup N_T\setminus \mathcal{J}\right\vert.
        \end{align*}

        For any $t\in \mathcal{J}$,
        we rewrite the hypothesis updating as follows.

        \begin{align*}
            \tilde{f}_{t+1}=&\bar{f}_t+\lambda y_t\kappa\left(\mathbf{x}_t,\cdot\right),\\
            f_{t+1}=&\mathop{\arg\min}_{f\in\mathbb{H}}
            \left\Vert f- \tilde{f}_{t+1}\right\Vert^2_{\mathcal{H}}.
        \end{align*}

        We follow the analysis of \eqref{eq:Supp_ECML2023:analysis_instantaneous_regret},
        but give some critical differences.

        \begin{align*}
                &\ell_{\mathrm{Hinge}}\left(f_t\left({\bf x}_t\right),y_t\right) -
                 \ell_{\mathrm{Hinge}}\left(f\left({\bf x}_t\right),y_t\right)\nonumber\\
            \leq&\left\langle \nabla_t,f_t-f \right\rangle\nonumber\\
            =   &\frac{1}{\lambda}\left\langle \bar{f}_t-\tilde{f}_{t+1},f_t-f \right\rangle\nonumber\\
            =   &\frac{1}{\lambda}\left\langle f_t-\tilde{f}_{t+1},f_t-f \right\rangle +
                 \frac{1}{\lambda}\left\langle \bar{f}_t-f_t,f_t-f \right\rangle\nonumber\\
            \leq&\frac{\left\Vert f_t-f\right\Vert^2_{\mathcal{H}} -
                 \left\Vert \tilde{f}_{t+1}-f \right\Vert^2_{\mathcal{H}}}{2\lambda} +
                 \frac{\left\Vert f_t-\tilde{f}_{t+1}\right\Vert^2_{\mathcal{H}}}{2\lambda}+\\
                &\frac{1}{\lambda}\left\langle \bar{f}_t-f_t,f_t-f \right\rangle\\
            =   &\frac{\left\Vert f_t-f \right\Vert^2_{\mathcal{H}} -
                 \left\Vert \tilde{f}_{t+1}-f \right\Vert^2_{\mathcal{H}}}{2\lambda} +
                 \frac{\left\Vert f_t-\bar{f}_t+\lambda\nabla_t \right\Vert^2_{\mathcal{H}}}{2\lambda}+\\
                &\frac{1}{\lambda}\left\langle \bar{f}_t-f_t,f_t-f \right\rangle\\
            =   &\frac{\left\Vert f_t-f \right\Vert^2_{\mathcal{H}} -
                 \left\Vert \tilde{f}_{t+1}-f \right\Vert^2_{\mathcal{H}}}{2\lambda} +
                 \frac{\lambda}{2}+\\
                &\frac{\left\Vert f_t-\bar{f}_t\right\Vert^2_{\mathcal{H}} +
                 2\left\langle\lambda\nabla_t,f_t-\bar{f}_t\right\rangle}{2\lambda} +
                 \frac{\left\langle \bar{f}_t-f_t,f_t-f \right\rangle}{\lambda}.
        \end{align*}

        Next we analyze
        \begin{align}
        \label{eq:supp_ECML2023:regret_removing_operation_1}
            &\frac{\left\Vert f_t-\bar{f}_t \right\Vert^2_{\mathcal{H}}+2\left\langle\lambda\nabla_t,f_t-\bar{f}_t \right\rangle}{2\lambda}+
             \frac{\left\langle \bar{f}_t-f_t,f_t-f \right\rangle}{\lambda}\nonumber\\
            =&\frac{\left\Vert \bar{f}_t-f \right\Vert^2_{\mathcal{H}}-\left\Vert f_t-f\right\Vert^2_{\mathcal{H}}}{2\lambda} +
              \left\langle\nabla_t,f_t-\bar{f}_t\right\rangle.
        \end{align}

        Summing over $t\in \mathcal{J}$ yields
        \begin{align*}
                &\sum_{t\in \mathcal{J}}\left[\ell_{\mathrm{Hinge}}\left(f_t\left({\bf x}_t\right),y_t\right) -
                                              \ell_{\mathrm{Hinge}}\left(f\left({\bf x}_t\right),y_t\right)\right]\\
            \leq&\sum_{t\in \mathcal{J}}\frac{\left\Vert f_t-f \right\Vert^2_{\mathcal{H}} -
                 \left\Vert f_{t+1}-f \right\Vert^2_{\mathcal{H}}}{2\lambda} +
                 \frac{\lambda}{2}J +\\
                &\sum_{t\in \mathcal{J}}\left[
                                                \frac{\left\Vert \bar{f}_t-f \right\Vert^2_{\mathcal{H}}}
                                                     {2\lambda} +
                                                \left\langle \nabla_t,f_t-\bar{f}_t \right\rangle
                                        \right].
        \end{align*}

        Finally,
        summing over $t\in M'_T\cup N_T$ gives
        \begin{align*}
                & \sum_{t\in M'_T\cup N_T}\left[ \ell_{\mathrm{Hinge}}\left(f_t\left({\bf x}_t\right),y_t\right) -
                                                 \ell_{\mathrm{Hinge}}\left(f\left({\bf x}_t \right),y_t\right)
                                          \right]\\
            \leq&\sum_{t\in M'_T\cup N_T}\frac{\left\Vert f_t-f \right\Vert^2_{\mathcal{H}} -
                                               \left\Vert f_{t+1}-f \right\Vert^2_{\mathcal{H}}}
                                              {2\lambda} +
                                         \lambda
                                         \frac{\left\vert M'_T\right\vert + \left\vert N_T \right\vert}{2} +\\
                &\sum_{t\in \mathcal{J}}\left[
                                                 \frac{\left\Vert \bar{f}_t-f \right\Vert^2_{\mathcal{H}}}
                                                      {2\lambda} +
                                                 \left\langle \nabla_t,f_t-\bar{f}_t \right\rangle
                                        \right]\\
            \leq&\frac{\left\Vert f \right\Vert^2_{\mathcal{H}}}{2\lambda} +
                 \frac{\lambda}{2}\left(\left\vert M'_T\right\vert +\left\vert N_T \right\vert\right) +\\
                &\left(\frac{2.56U^2}{2\lambda} + 1.6U \right)
                 \left(\frac{2\left(\left\vert M'_T \right\vert + \left\vert N_T \right\vert\right)}
                            {B} - 1\right).
        \end{align*}

        Let $\lambda=\dfrac{\sqrt{2}U}{\sqrt{B}}$, $B\geq 50$.

        According to \eqref{eq:Supp_ECML2023:gap:losses_and_mistakes},
        the mistake bound of Ahpatron satisfies
        \begin{align*}
                &\left\vert M'_T \right\vert -L_T(f)\\
            \leq&\sum_{t\in M'_T\cup N_T}\ell_{\mathrm{Hinge}}\left(f_t\left({\bf x}_t\right),y_t\right) -
                                         \ell_{\mathrm{Hinge}}\left(f\left({\bf x}_t\right),y_t\right) -
                                         \varepsilon \left\vert N_T \right\vert\\
            \leq&\frac{\Vert f\Vert^2_{\mathcal{H}}}{2\lambda} +
                 \frac{\lambda}{2}\left(\left\vert M'_T \right\vert + \left\vert N_T \right\vert \right)+\\
                &\left(\frac{2.56U^2}{2\lambda} + 1.6U \right)
                 \left(\frac{2\left(\left\vert M'_T \right\vert + \left\vert N_T \right\vert\right)}{B} -1 \right) -
                 \varepsilon \left\vert N_T \right\vert\\
            \leq&\left(\frac{\lambda}{2} + \frac{2.56U^2}{\lambda B} + \frac{3.2U}{B} \right)\left\vert M'_T\right\vert +\\
                &\left(\frac{\lambda}{2} + \frac{2.56U^2}{\lambda B} + \frac{3.2U}{B}- \varepsilon \right) \left\vert N_T\right\vert\\
            \leq&\frac{3U}{\sqrt{B}}\left\vert M'_T \right\vert +
                 \left(\frac{3U}{\sqrt{B}} - \varepsilon \right)\left\vert N_T \right\vert.
        \end{align*}
        Let
        $$
            1-\frac{3U}{\sqrt{B}}\geq \frac{1}{4},\quad
            \frac{3U}{\sqrt{B}}< \varepsilon <1.
        $$
        We obtain $U\leq \dfrac{\sqrt{B}}{4}$ and
        \begin{align*}
                &\left\vert M'_T\right\vert -L_T(f)\\
            \leq&\frac{3U}{\sqrt{B}} \frac{L_T(f)}{1-\frac{3U}{\sqrt{B}}} +
                 \frac{\frac{3U}{\sqrt{B}} - \varepsilon} {1-\frac{3U}{\sqrt{B}}}\left\vert N_T\right\vert\\
            \leq&\frac{12U}{\sqrt{B}}L_T(f) +
                 \frac{\frac{3U}{\sqrt{B}} - \varepsilon} {1-\frac{3U}{\sqrt{B}}}\left\vert N_T\right\vert.
        \end{align*}

        Next we consider the case $J=0$.

        \begin{align*}
                &\left\vert M'_T\right\vert - L_T(f)\\
            \leq&\frac{\left\Vert f\right\Vert^2_{\mathcal{H}}}{2\lambda} +
                 \lambda\frac{\left\vert M'_T\right\vert}{2} +
                 \left(\frac{\lambda}{2} - \varepsilon \right)\left\vert N_T\right\vert\\
            \leq&\frac{\left\Vert f\right\Vert^2_{\mathcal{H}}\sqrt{B}}{2\sqrt{2}U} +
                 \frac{U \left\vert M'_T \right \vert}{\sqrt{2B}} +
                 \left(\frac{\lambda}{2}-\varepsilon\right)\left\vert N_T\right\vert.
        \end{align*}
        Using the fact $U\leq\dfrac{\sqrt{B}}{4}$ and rearranging terms yields
        \begin{align*}
                &\left\vert M'_T \right\vert - L_T(f)\\
            \leq&\frac{\frac{U}{\sqrt{2B}}}{1-\frac{U}{\sqrt{2B}}}L_T(f) +
                 \frac{\sqrt{B}}{1-\frac{U}{\sqrt{2B}}} \frac{\Vert f\Vert^2_{\mathcal{H}}}{2\sqrt{2}U} +
                 \frac{\frac{U}{\sqrt{2B}}-\varepsilon}{1-\frac{U}{\sqrt{2B}}}\left\vert N_T\right\vert\\
            \leq&\frac{0.9U}{\sqrt{B}}L_T(f) +
                 \frac{\sqrt{B}}{2U}\Vert f\Vert^2_{\mathcal{H}} +
                 \left(\frac{U}{\sqrt{2B}} - \varepsilon\right) \frac{\left\vert N_T \right \vert} {1-\frac{U}{\sqrt{2B}}}.
        \end{align*}
        Combining the two cases
        and using the fact $\vert M_T\vert \leq \vert M'_T\vert$ concludes the proof.
    \end{proof}

\section{Proof of Theorem \ref{eq:AAAI2024:kernel_alignment:small_loss}}

    Let
    $$
    \bar{f}=\dfrac{1}{T}\sum^T_{t=1}y_t\kappa\left({\bf x}_t,\cdot\right).
    $$
    For any ${\bf x}\in\mathcal{X}$, it can be verified that
    $
        \left\Vert\bar{f} \right\Vert_{\mathcal{H}} \leq 1
    $ and
    $
        \bar{f}\left({\bf x}\right) \leq 1
    $.

    Let $U\geq 1$, then $\bar{f}\in\mathbb{H}$.
    Thus we can obtain
    \begin{align*}
        \min_{f\in\mathbb{H}}L_T(f)
            &\leq L_T\left(\bar{f}\right)\\
            &=\sum^T_{t=1}\left(1-y_t\bar{f}\left({\bf x}_t\right)\right)\\
            &=T - \frac{1}{T} \sum^T_{t=1}\sum^T_{\tau=1}y_t y_\tau\kappa\left({\bf x}_t,{\bf x}_\tau\right)\\
            &=\mathcal{A}_T,
    \end{align*}
    which concludes the proof.

\section{Proof of Theorem \ref{thm:AAAI24:refined_mistake_bound_of_H_Forgetron}}

    \begin{proof}
        The overall analysis is similar to that of Theorem 4.
        According to \eqref{eq:supp_ECML2023:regret_removing_operation_1},
        the mistake bound depends on the distance between $\bar{f}_t$ and $f_t$ via the following term.
        \begin{align*}
                &\sum_{t\in \mathcal{J}} \frac{\left\Vert f_t- \bar{f}_t \right\Vert^2_{\mathcal{H}} +
                                               2\left\langle \lambda\nabla_t, f_t - \bar{f}_t\right\rangle}
                                              {2\lambda} +
                                         \frac{\left\langle \bar{f}_t - f_t, f_t-f \right\rangle}
                                              {\lambda}\\
            =   &\sum_{t\in \mathcal{J}} \left[\frac{\left\langle f_t-\bar{f}_t, f_t-\bar{f_t}+ 2\lambda\nabla_t - 2f_t+2f\right\rangle}
                                                    {2\lambda}
                                         \right]\\
            \mathop{=}^{(a)}
                &\sum_{t\in \mathcal{J}} \frac{\left\Vert \bar{f}_t \right\Vert^2_{\mathcal{H}} -\left\Vert f_t \right\Vert^2_{\mathcal{H}}}
                                              {2\lambda} +
                 \sum_{t\in \mathcal{J}} \frac{\left\langle f_t-\bar{f}_t,2\lambda\nabla_t+2f \right\rangle}
                                              {2\lambda}\\
            \leq&\frac{(U+\lambda)\sum_{t\in \mathcal{J}}\left\Vert f_t-\bar{f}_t\right\Vert_{\mathcal{H}}}
                      {\lambda}\\
            \mathop{\leq}^{(b)}
                &\frac{U+\lambda}{\lambda} \zeta UJ.
        \end{align*}

        Let $c_t=\dfrac{\left\Vert f_t\right\Vert_{\mathcal{H}}}{U}$.\\
        In equation $(a)$,
        we have $\left\Vert\bar{f}_t\right\Vert^2_{\mathcal{H}} = \left\Vert f_t\right\Vert^2_{\mathcal{H}}$.

        There must be a constant $\zeta\in(0,2]$ such that
        the inequality $(b)$ holds.
        In the worst-case, $\zeta=2$.
        If $\bar{f}_t$ is close to $f_t$ for most of $t\in\mathcal{J}$,
        then $\zeta\ll 2$ is possible.

        Let $\lambda=\dfrac{U}{2\sqrt{B}}$ and $B\geq 16$.
        We first consider the case $J\geq 1$.

        According to \eqref{eq:Supp_ECML2023:gap:losses_and_mistakes},
        the mistake bound of Ahpatron satisfies
        \begin{align*}
                &\left\vert M'_T\right\vert - L_T(f)\\
            \leq&\sum_{t\in M'_T\cup N_T} \frac{\left\Vert f_t-f\right\Vert^2_{\mathcal{H}} -
                                                \left\Vert f_{t+1}-f\right\Vert^2_{\mathcal{H}}}
                                               {2\lambda} +
                                          \frac{\left\vert M'_T\right\vert + \left\vert N_T\right\vert}
                                               {2}
                                          \lambda +\\
                &\frac{U+\lambda}{\lambda} \zeta UJ - \varepsilon \left\vert N_T\right\vert\\
            \leq&\frac{\Vert f\Vert^2_{\mathcal{H}}}
                      {2\lambda} +\frac{\lambda}{2}\left(\left\vert M'_T\right\vert + \left\vert N_T\right\vert\right) -
                 \varepsilon \left\vert N_T\right\vert+\\
                &\frac{U+\lambda}{\lambda} \zeta U \left(\frac{2\left(\left\vert M'_T\right\vert + \left\vert N_T\right\vert\right)}
                                                              {B} -
                                                         1
                                                   \right)\\
            \leq&\frac{\Vert f\Vert^2_{\mathcal{H}}}{2\lambda} +
                 \left(\frac{\lambda}{2} + \frac{2U^2\zeta}{B\lambda} + \frac{2\zeta U}{B} \right)\left\vert M'_T\right\vert -
                 \frac{U^2\zeta}{\lambda}+\\
                &\left(\frac{\lambda}{2}+\frac{2U^2\zeta}{B\lambda}+\frac{2\zeta U}{B} -\varepsilon \right)\left\vert N_T\right\vert\\
            \leq&(1-2\zeta)\frac{\Vert f\Vert^2_{\mathcal{H}}\sqrt{B}}{U}+
                 \left(\frac{1}{4}+\frac{9\zeta}{2}\right)\frac{U}{\sqrt{B}}\left\vert M'_T\right\vert+\\
                &\left(\left(\frac{1}{4}+\frac{9\zeta}{2}\right)\frac{U}{\sqrt{B}}-\varepsilon\right)\left\vert N_T\right\vert,
        \end{align*}
        where we require
        $$
            \left(\frac{1}{4}+\frac{9\zeta}{2}\right)\frac{U}{\sqrt{B}}<\varepsilon<1.
        $$
        Rearranging terms yields
        \begin{align*}
                &\left\vert M'_T\right\vert\\
            \leq&\frac{L_T(f)}{1-\left(\frac{1}{4}+\frac{9\zeta}{2}\right)\frac{U}{\sqrt{B}}} +
                 \frac{(1-2\zeta)\frac{\Vert f\Vert^2_{\mathcal{H}}\sqrt{B}}{U}}
                      {1-\left(\frac{1}{4}+\frac{9\zeta}{2}\right)\frac{U}{\sqrt{B}}}+\\
                &\underbrace{\left(\left(\frac{1}{4}+\frac{9\zeta}{2}\right)\frac{U}{\sqrt{B}}-\varepsilon\right)
                                         \frac{1}
                                              {1-\left(\frac{1}{4}+\frac{9\zeta}{2}\right)\frac{U}{\sqrt{B}}}
                                         \left\vert N_T\right\vert}_{\Delta_1}.
        \end{align*}
        For any $\gamma\in(0,1)$,
        let $U\leq\dfrac{(1-\gamma)\sqrt{B}}{\frac{1}{4}+\frac{9}{2}\zeta}$.

        We further obtain
        \begin{equation}
        \label{eq:supp_ECML2023:alg_dep_mistake_bound_H_Forgetron:case_one}
        \begin{split}
                &\left\vert M'_T\right\vert - L_T(f)\\
            \leq&\frac{1}{\gamma}\left(\frac{1}{4}+\frac{9\zeta}{2}\right)\frac{UL_T(f)}{\sqrt{B}} +
                 \frac{1-2\zeta}{\gamma}\frac{\Vert f\Vert^2_{\mathcal{H}}\sqrt{B}}{U} +
                 \Delta_1.
        \end{split}
        \end{equation}

        Next we consider the case $J=0$.\\

        \begin{align*}
                &\left\vert M'_T\right\vert -L_T(f)\\
            \leq
                &\frac{\Vert f\Vert^2_{\mathcal{H}}}{2\lambda} +
                 \lambda\frac{\left\vert M_T\right\vert}{2} +
                 \left(\frac{\lambda}{2}-\varepsilon\right)\left\vert N_T\right\vert\\
            \leq
                &\frac{\Vert f\Vert^2_{\mathcal{H}}\sqrt{B}}{U} +
                 \frac{U\left\vert M_T \right\vert}{4\sqrt{B}} +
                 \left(\frac{\lambda}{2}-\varepsilon\right)\left\vert N_T\right\vert.
        \end{align*}

        Let $U\leq 4(1-\gamma)\sqrt{B}$.\\
        Rearranging terms yields
        \begin{align*}
                &\left\vert M'_T\right\vert -L_T(f)\\
            \leq&\frac{\frac{U}{4\sqrt{B}}}{1-\frac{U}{4\sqrt{B}}}L_T(f)+
                 \frac{\sqrt{B}}{1-\frac{U}{4\sqrt{B}}}\frac{\Vert f\Vert^2_{\mathcal{H}}}{U} +
                 \frac{\frac{U}{4\sqrt{B}}-\varepsilon}{1-\frac{U}{4\sqrt{B}}}\left\vert N_T\right\vert\\
            \leq&\frac{U}{4\gamma\sqrt{B}}L_T(f)+
                 \frac{1}{\gamma}\frac{\Vert f\Vert^2_{\mathcal{H}}\sqrt{B}}{U} +
                 \left(\frac{U}{4\sqrt{B}}-\varepsilon\right)\frac{\left\vert N_T\right\vert}{1-\frac{U}{4\sqrt{B}}}.
        \end{align*}
        The mistake bound is smaller than \eqref{eq:supp_ECML2023:alg_dep_mistake_bound_H_Forgetron:case_one}.

        Combining the two cases
        and using the fact $\vert M_T\vert \leq \vert M'_T\vert$ concludes the proof.
    \end{proof}
\section{More Experiments}

    We further compare Ahpatron with Projecton++.
    We will increase the budget used by Ahpatron.
    The goal is to show that
    Ahpatron performs better than Projecton++ on the same or a smaller budget.
    We do not change the configuration of parameters of the two algorithms.
    The results are shown in Table \ref{tab:JMLR2022:memory_mistakes:further_compare_with_projectron}.
\begin{table}[!hbt]
      \centering
      \setlength{\tabcolsep}{1.3mm}
      \caption{
      Further comparison with Projectron++}
      \label{tab:JMLR2022:memory_mistakes:further_compare_with_projectron}
      \begin{tabular}{l|rrr}
        \toprule
        \multirow{2}{*}{Algorithm}&\multicolumn{3}{c}{SUSY, $\sigma=1$}\\
        \cline{2-4}&AMR (\%) &$B$       &Time (s)  \\
        \hline
        Projectron++         & 22.05 $\pm$ 0.07 & 4898 & 1330.83\\
        Ahpatron          & \textbf{21.97} $\pm$ \textbf{0.11} & 3000 & 38.20    \\
        \midrule
        \multirow{2}{*}{Algorithm}&\multicolumn{3}{c}{ijcnn1, $\sigma=1$}\\
        \cline{2-4}&AMR (\%) &$B$       &Time (s)  \\
        \hline
        Projectron++      & \textbf{3.01} $\pm$ \textbf{0.04} & 1577  & 58.58\\
        Ahpatron          & 3.14 $\pm$ 0.03 & 1570  & 22.87    \\
        \midrule
        \multirow{2}{*}{Algorithm}&\multicolumn{3}{c}{cod-rna, $\sigma=1$}\\
        \cline{2-4}&AMR (\%) &$B$       &Time (s)  \\
        \hline
        Projectron++         & 10.13 $\pm$ 0.03 & 5100  & 2786.98\\
        Ahpatron          & \textbf{10.09} $\pm$ \textbf{0.03} & 3000 & 204.08    \\
        \midrule
        \multirow{2}{*}{Algorithm}&\multicolumn{3}{c}{w8a, $\sigma=2$}\\
        \cline{2-4}&AMR (\%) &$B$       &Time (s)  \\
        \hline
        Projectron++         & 2.03  $\pm$ 0.03 & 1279  & 94.37\\
        Ahpatron          & \textbf{1.93}  $\pm$ \textbf{0.05} & 500 & 16.83    \\
        \bottomrule
      \end{tabular}
    \end{table}

    Most of the results verify that
    Ahpatron has smaller average mistake rates (AMR) than Projetron++
    on the same or a smaller budget.
    Besides,
    the running time of Ahpatron is also smaller.

\end{document}